\PassOptionsToPackage{table}{xcolor}
\documentclass[sigconf]{acmart}

\newif\ifdraft
\draftfalse

\ifdraft

\else

\fi

\usepackage{color}
\usepackage{xspace}
\usepackage{xcolor}

\makeatletter
\DeclareRobustCommand\onedot{\futurelet\@let@token\@onedot}
\def\@onedot{\ifx\@let@token.\else.\null\fi\xspace}

\def\eg{\emph{e.g}\onedot}

\def\etc{\emph{etc}\onedot}

\def\etal{\emph{et al}\onedot}
\makeatother

\newcommand{\figref}[1]{Figure~\ref{fig:#1}}%
\newcommand{\tabref}[1]{Table~\ref{tab:#1}} %

\newcommand{\secref}[1]{Section~\ref{sec:#1}}

\newlength{\qualw}     %
\newcommand{\image}{\mathbf{x}}%
\newcommand{\matte}{\boldsymbol{\alpha}}%
\newcommand{\latent}{\mathbf{z}}%
\newcommand{\noise}{\boldsymbol{\epsilon}}%

\definecolor{cvprblue}{rgb}{0.21,0.49,0.74}
\definecolor{darkgreen}{rgb}{0.00,0.40,0.0.00}

\colorlet{dark-blue}{blue!50!black}
\colorlet{dark-cyan}{cyan!75!black}
\colorlet{dark-purple}{purple!50!black}
\colorlet{dark-red}{red!75!black}
\colorlet{dark-green}{green!80!black}
\colorlet{dark-orange}{orange!50!black}
\colorlet{dark-gray}{black!75}
\colorlet{light-gray}{black!30}
\definecolor{nice-red}{HTML}{E41A1C}
\definecolor{nice-orange}{HTML}{FF7F00}
\definecolor{nice-yellow}{HTML}{FFC020}
\definecolor{nice-green}{HTML}{39b54a}
\definecolor{nice-blue}{HTML}{0071bc}
\definecolor{nice-purple}{HTML}{984EA3}

\graphicspath{{figure/}}

    \makeatletter
\def\@fnsymbol#1{\ensuremath{\ifcase#1\or \dagger\or \ddagger\or
   \mathsection\or \mathparagraph\or \|\or **\or \dagger\dagger
   \or \ddagger\ddagger \else\@ctrerr\fi}}
    \makeatother

\copyrightyear{2024}
\acmYear{2024}
\setcopyright{rightsretained}
\acmConference[SIGGRAPH Conference Papers '24]{Special Interest Group on Computer Graphics and Interactive Techniques Conference Conference Papers '24}{July 27-August 1, 2024}{Denver, CO, USA}
\acmBooktitle{Special Interest Group on Computer Graphics and Interactive Techniques Conference Conference Papers '24 (SIGGRAPH Conference Papers '24), July 27-August 1, 2024, Denver, CO, USA}
\acmDOI{10.1145/3641519.3657519}
\acmISBN{979-8-4007-0525-0/24/07}

\begin{document}

\title{Matting by Generation
}

\settopmatter{authorsperrow=4}

\author[Wang]{Zhixiang Wang}
\orcid{0000-0002-5016-587X}
\affiliation{%
  \institution{The University of Tokyo}
  \city{Tokyo}
  \country{Japan}
}
\email{wangzx@nii.ac.jp}

\author[Li]{Baiang Li}
\orcid{0009-0005-3844-2554}
\affiliation{%
  \institution{Hefei University of Technology}
  \city{Hefei}
  \country{China}}
\email{ztmotalee@gmail.com}

\author[Wang]{Jian Wang}
\orcid{0000-0001-5266-3808}
\authornote{Corresponding author}
\affiliation{%
  \institution{Snap Research}
  \city{New York}
  \country{USA}
}
\email{jwang4@snap.com}

\author[Liu]{Yu-Lun Liu}
\orcid{0000-0002-7561-6884}
\affiliation{%
 \institution{National Yang Ming Chiao Tung University}
 \city{Hsinchu}
 \country{Taiwan}}
 \email{yulunliu@cs.nycu.edu.tw}

\author[Gu]{Jinwei Gu}
\orcid{0000-0001-8705-8237}
\affiliation{%
  \institution{The Chinese University of Hong Kong}
  \city{}
  \country{Hong Kong SAR}}
\email{jwgu@cuhk.edu.hk}

\author[Chuang]{Yung-Yu Chuang}
\orcid{0000-0002-1383-0017}
\affiliation{%
  \institution{National Taiwan University}
  \city{Taipei}
  \country{Taiwan}}
\email{cyy@csie.ntu.edu.tw}

\author[Satoh]{Shin'ichi Satoh}
\orcid{0000-0001-6995-6447}
\affiliation{%
  \institution{National Institute of Informatics}
  \city{Tokyo}
  \country{Japan}}
\email{satoh@nii.ac.jp}

\citestyle{acmauthoryear}

\begin{teaserfigure}
\small
\centering
\newlength{\teaserw}
\setlength{\teaserw}{0.245\linewidth}
\newcommand{\teaserrowup}[5]{
 {\includegraphics[trim={#2 #3 #4 #5}, clip,width=\teaserw]{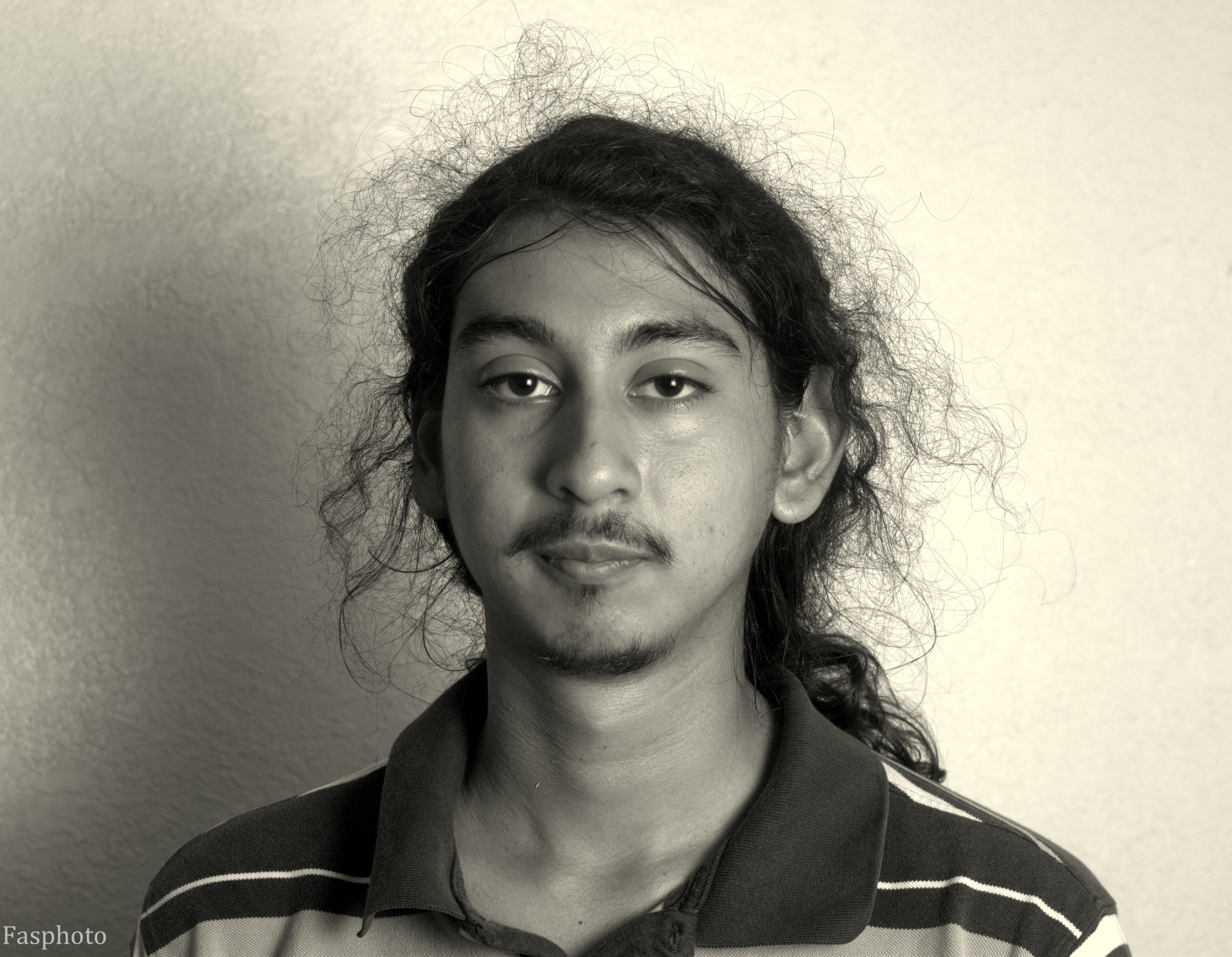}} & 
 {\includegraphics[trim={#2 #3 #4 #5}, clip,width=\teaserw]{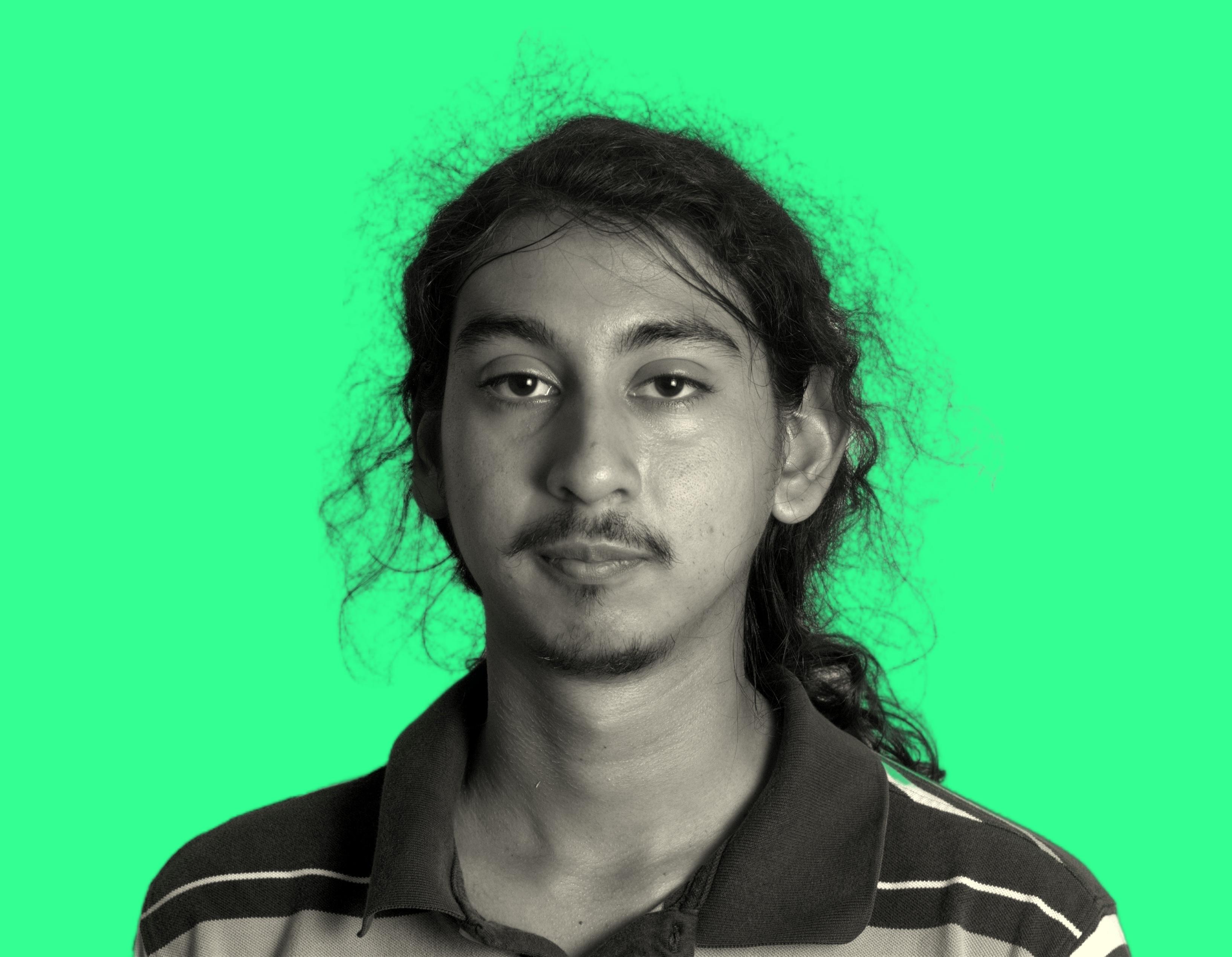}} & 
  {\includegraphics[trim={#2 #3 #4 #5}, clip,width=\teaserw]{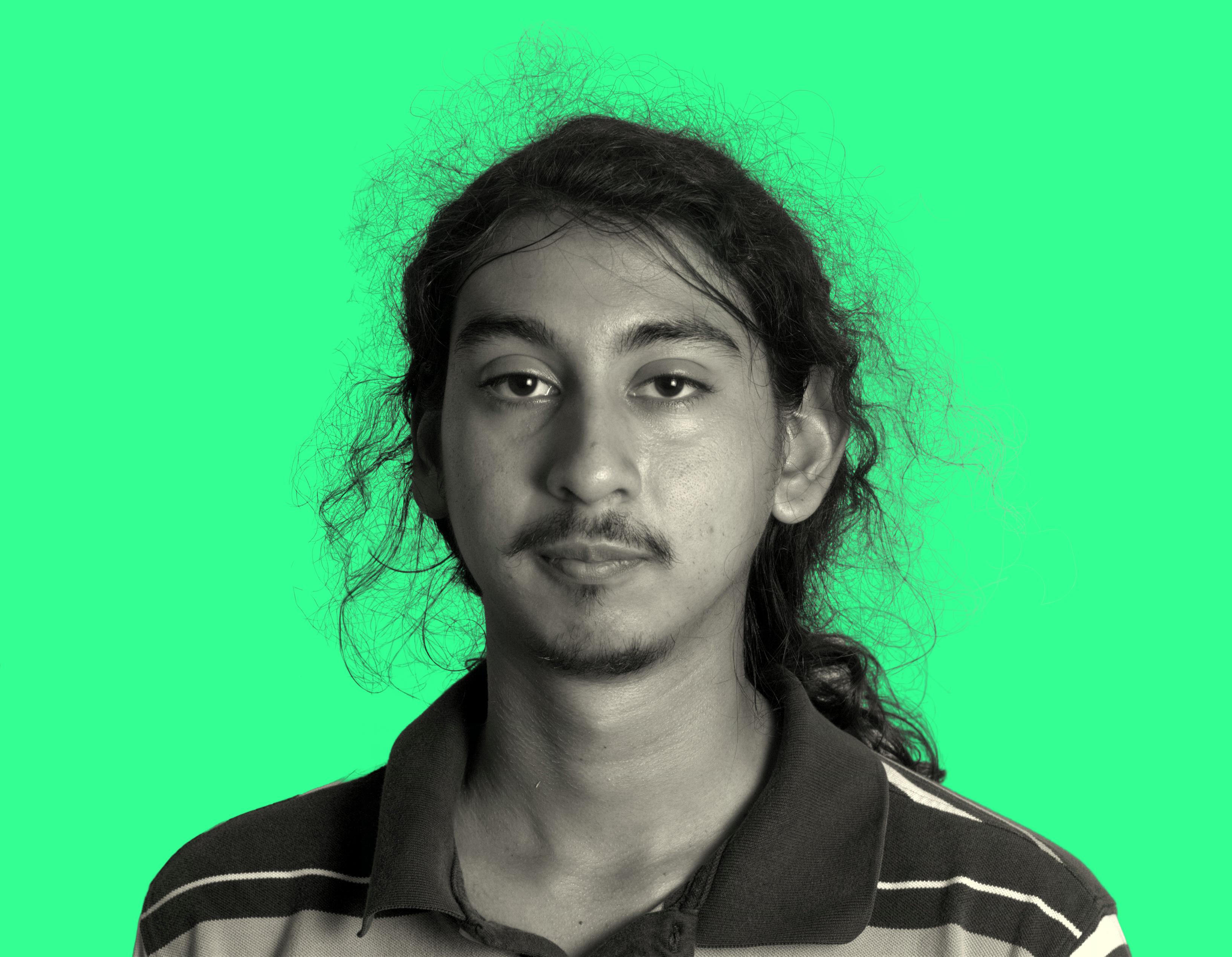}} &
  {\includegraphics[trim={#2 #3 #4 #5}, clip,width=\teaserw]{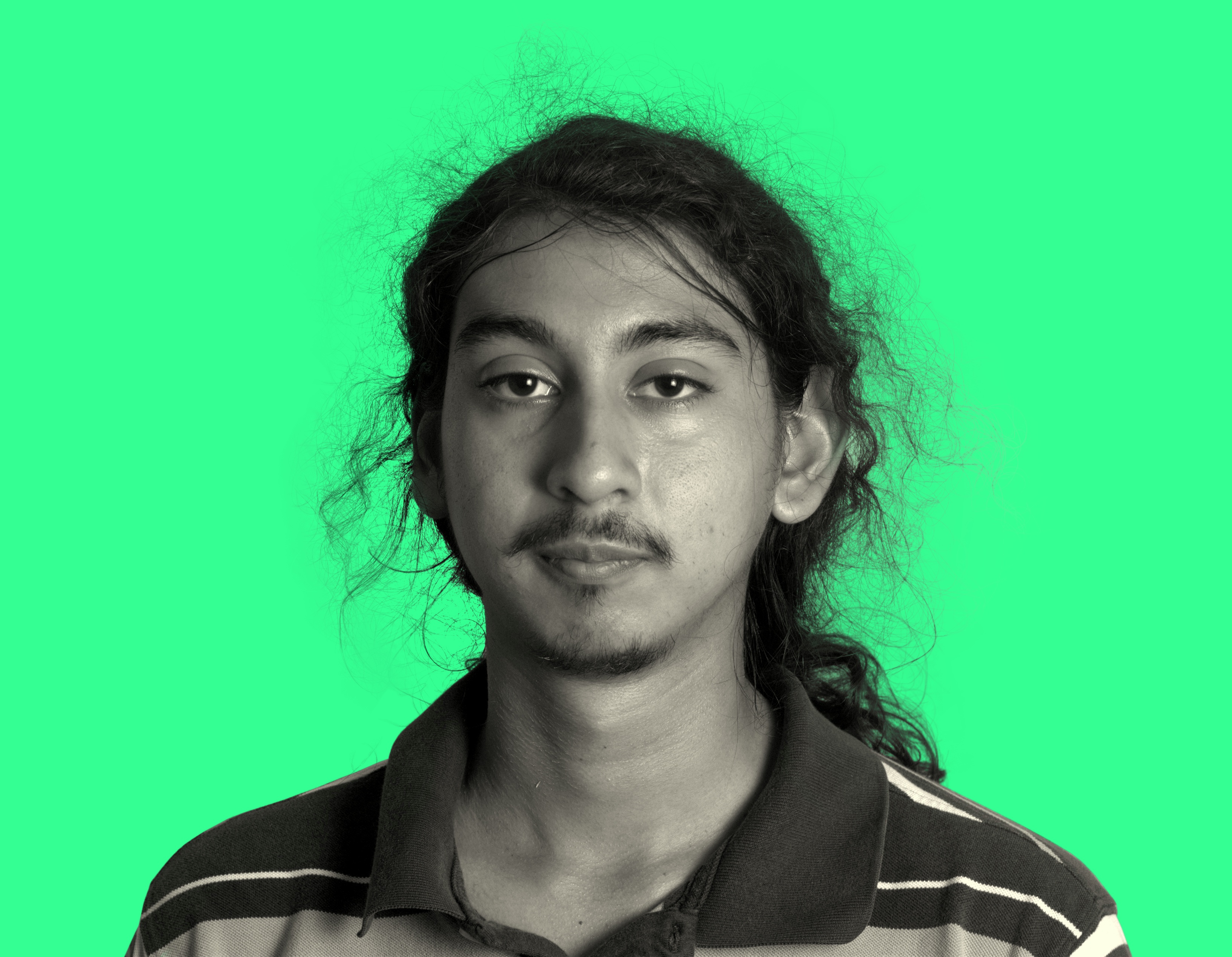}}
}
\newcommand{\teaserrowbottom}[5]{
{\includegraphics[trim={#2 #3 #4 #5}, clip,width=\teaserw]{teaser/#1.jpg}} &
{\includegraphics[trim={#2 #3 #4 #5}, clip,width=\teaserw]{teaser/#1_p3m-net.jpg}} &
{\includegraphics[trim={#2 #3 #4 #5}, clip,width=\teaserw]{teaser/#1_ours.jpg}} &
{\includegraphics[trim={#2 #3 #4 #5}, clip,width=\teaserw]{teaser/#1_gt.jpg}}
 \\
 {Input} & ViTAE-S~\cite{MA-2023-P3M-ViTAE} &  Ours & Human annotation\\
}
\small
\begin{tabular}
    {   
        @{\hspace{0mm}}c@{\hspace{0.8mm}} 
        @{\hspace{0mm}}c@{\hspace{0.8mm}}
        @{\hspace{0mm}}c@{\hspace{0.8mm}}
        @{\hspace{0mm}}c@{\hspace{0.8mm}} 
    }
    \centering
    \teaserrowup{input_patch}{0}{0}{0}{0}\\
    \teaserrowbottom{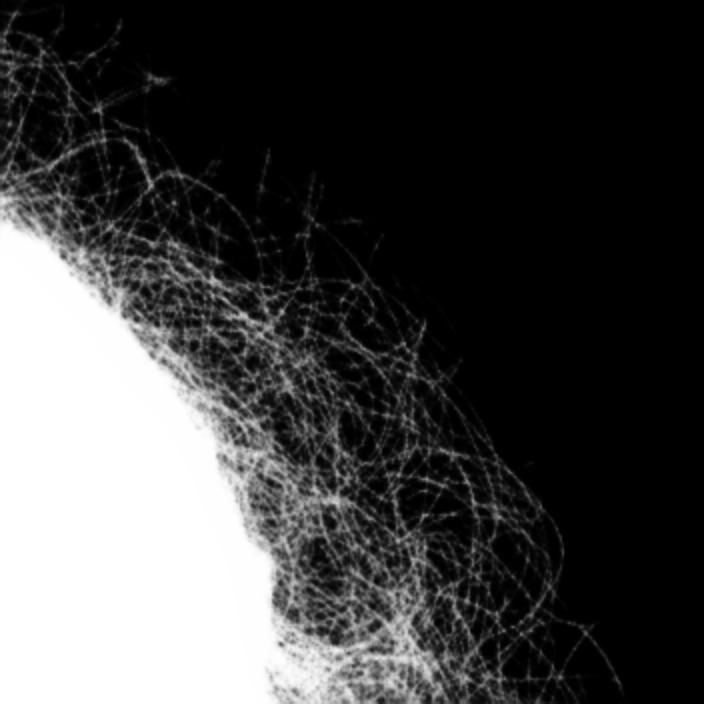}{0}{0}{0}{0}\\
\end{tabular}
\caption{
Matting by Generation. We crack the trimap-free matting problem in a conditional {generative} way as opposed to the previous regression-based fashion. With only an image as input, our method automatically extracts the foreground (\eg, person) and generates high-quality boundary details benefiting from the rich generative prior, leading to photorealistic compositions. 
Compared with the human annotation, our results provide crisper details and greater fidelity to the input image in this example. 
}
    \label{fig:teaser}
    \Description{Matting by Generation.}
\end{teaserfigure}

\begin{abstract}
This paper introduces an innovative approach for image matting that redefines the traditional regression-based task as a generative modeling challenge. Our method harnesses the capabilities of latent diffusion models, enriched with extensive pre-trained knowledge, to regularize the matting process. We present novel architectural innovations that empower our model to produce mattes with superior resolution and detail. 
The proposed method is versatile and can perform both guidance-free and guidance-based image matting, accommodating a variety of additional cues.
Our comprehensive evaluation across three benchmark datasets demonstrates the superior performance of our approach, both quantitatively and qualitatively. The results not only reflect our method's robust effectiveness but also highlight its ability to generate visually compelling mattes that approach photorealistic quality. 
The project page for this paper is at \url{https://lightchaserx.github.io/matting-by-generation/}.
\end{abstract}



\maketitle

\section{Introduction}
\label{sec:intro}

Image matting as a fundamental problem in computer vision has been investigated for decades~\cite{zhang2020empowering}. It enables many real applications, such as visual effects synthesis \cite{li2022ganimator}, image editing \cite{kawar2023imagic}, \etc.
Its goal is to predict the foreground and the alpha matte from an input image.
This is a highly ill-posed inverse problem with only the input being known. The forward model is the composition equation~\cite{porter1984compositing} given by:
\begin{equation}
    C = \alpha F + \left(1 - \alpha\right) B\,,
\end{equation}
where $C$ is the input, $F$ is the foreground, $B$ is the background, and $\alpha\in[0,1]$ is the linear combination coefficient. 
The main challenge lies in the ill-posedness, which is a mixed difficulty --- to find where the foreground is and what the opacity value is in the boundary.

Existing methods, regardless of traditional or learning-based, leverage additional inputs to reduce the ill-posedness.
For example, one could mitigate unknown parameter $B$ by capturing another background image~\cite{sengupta2020background,lin2021real},
or could add priors for $\alpha$ through user annotated trimaps.\footnote{Even with a known background image $B$, the problem is still ill-posed (3 unknowns for foreground color, plus unknown alpha, for 3 equations across R, G, B channels.)} 
Besides using additional input provided by humans, methods~\cite{Yu-2021-MG,Li-2023-MattingAny} employing rough masks from other algorithms, such as Segment Anything (SAM)~\cite{SAM}, aim to alleviate the training burden in segmentation and enhance focus on boundary matte quality. However, these approaches are not entirely satisfactory, primarily due to their reliance on the segmentation network's accuracy. Imprecise initial segmentation can significantly compromise the quality of matting results, particularly at the boundaries. This dependency raises concerns about the efficacy of solely relying on rough segmentation masks for achieving high-quality matting.

Recent advancements in end-to-end matting methods~\cite{Li-2021-P3M,Ke-2022-MODNet} have sought to address these limitations by eliminating the need for these additional inputs, thereby reducing the reliance on human-generated data. Nevertheless, developing an effective end-to-end matting algorithm from scratch poses significant challenges due to the task's inherent complexity. 
These methods typically employ strategies such as constraining the application domain to portrait images~\cite{Li-2021-P3M, MA-2023-P3M-ViTAE} and imposing implicit segmentation prior~\cite{Ke-2022-MODNet}.
While these approaches reduce ambiguity between segmentation and matting and encourage the model to capture boundary details more effectively, achieving high-quality boundary mattes remains challenging, as shown in Figure~\ref{fig:teaser}. 
The prevailing issue with existing matting approaches lies in their handling of boundary regions, which are often challenging due to factors such as \emph{low visibility} (contrast, image quality) and \emph{imperfect} human annotations\footnote{Although there are solutions for capturing ground-truth alpha matte~\cite{smirnov2023magenta}, they often involve hardware requirements and are hard to scale up.}. These limitations can result in unnatural compositions, highlighting the need for more sophisticated solutions.

\begin{figure}
\small
\centering
\begin{tabular}{@{\hspace{0mm}}c@{\hspace{1.2mm}} @{\hspace{0mm}}c@{\hspace{1.2mm}}@{\hspace{0mm}}c@{\hspace{1.2mm}}@{\hspace{0mm}}c@{\hspace{1.2mm}}}
\includegraphics[width=0.24\columnwidth]{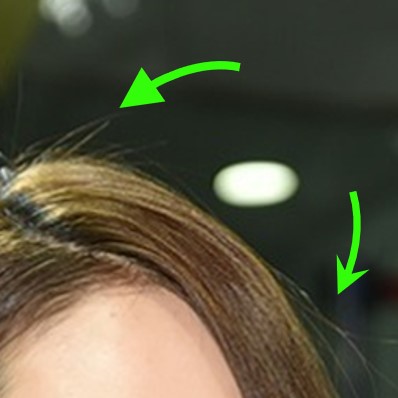} & 
\includegraphics[width=0.24\columnwidth]{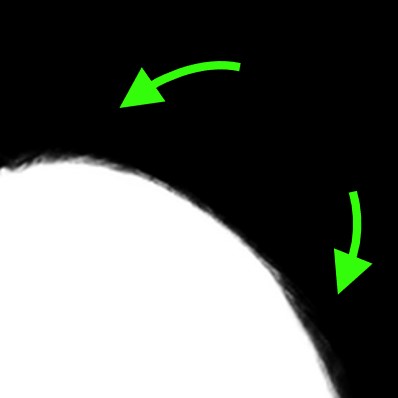} &  
\includegraphics[width=0.24\columnwidth]{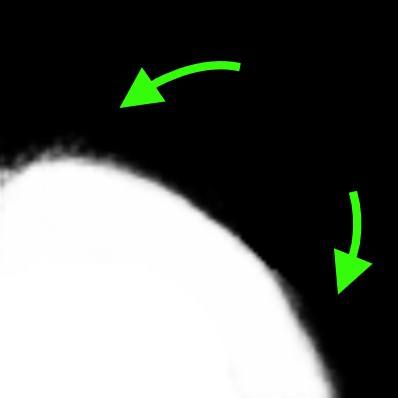} & 
\includegraphics[width=0.24\columnwidth]{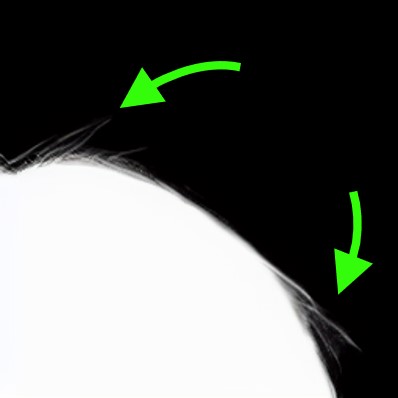}\\
Training sample & Annotation & ViTAE-S & Ours\\
\end{tabular}
\caption{Imperfect human annotation.
The training data are usually either blurry or lacking in some details. Therefore, the regression-based model would overfit the imperfect ground truth.}
\Description{Imperfect human annotation.}
\label{fig:imperfect_label}
\end{figure}

In this paper, we propose a simple yet effective technique of matting by generation. We transform the traditional \emph{regression} problem into a conditional \emph{generative} modeling problem, leveraging a diffusion model enriched with pre-trained knowledge about image semantics and matte details.
There are several key advantages to this approach.
Firstly, the generative model is adept at handling the uncertainties inherent in data, enabling it to learn the matte distribution more effectively than regression models. It also allows us to mitigate the negative impact of imperfect labels, such as ground-truth (GT) mattes generated by either humans or machines. These GT mattes, often derived from low-level image features, tend to contain imperfections, as exemplified in \figref{teaser}. Utilizing such flawed mattes to train a regression-based model can lead to overfitting and suboptimal outcomes, as demonstrated in \figref{imperfect_label}. In contrast, our generative prior empowers our method to identify semantically correct boundaries and even generate results surpassing the GT mattes' quality.
Secondly, our pre-trained diffusion model, with its vast database of billions of images, captures a more comprehensive image distribution. This broader understanding aids in regularizing the training process, offering more detailed and low-level property insights.
Thus, the generative capabilities of our model shine in scenarios where image visibility is limited.
Finally, our method offers versatility, accommodating both guidance-free and guidance-based matting. In most instances, it can perform accurate matting without additional hints. Nevertheless, in cases where the foreground is ambiguous, users can provide supplementary guidance to extract the desired matte.

\paragraph{List of Contributions}
Our research makes the following three significant contributions:
\begin{itemize}
    \item  We convert the regression problem into a generative modeling problem, utilizing generative diffusion prior to regularize training effectively.
    \item  We develop a model capable of processing high-resolution inputs efficiently and effectively.
    \item Our model is versatile and capable of handling scenarios with a variety of hints, including trimaps, masks, texts, and no hints at all.
\end{itemize}

\section{Related Work}
\label{sec:related}

\paragraph{Guidance-based Matting.}
Given only a single image, matting is an ill-posed inverse problem. Therefore, some matting methods require additional guidance, such as trimaps~\cite{chuang2001bayesian}, scribles~\cite{levin2008closed}, and clicks~\cite{wei2021click}. 
Methods of this category are typically referred to as guidance-based or trimap-based methods. 
Conventional trimap-based matting methods can be roughly divided into two categories: sampling-based methods~\cite{chuang2001bayesian, wang2007robust, gastal2010shared, he2011global, shahrian2013improving, feng2016cluster, yang2018active} and propagation-based methods~\cite{sun2004poisson, Grady2005randomwalk, wang2007robust, levin2008closed, chen2012knn, aksoy2017infoflow}.
Sampling-based methods usually resolve matting on a pixel-by-pixel basis by collecting color samples and forming a probabilistic distribution for each pixel's neighborhood.
In contrast, propagation-based methods aim to obtain the matte for the entire image at once by establishing pixel affinities and solving an equation. 
Complex scenes often pose a challenge to these methods. In recent years, deep learning has been introduced to solve the matting problem and gained success~\cite{cho2016init, xu2017deep, lu2019indices, hou2019context, sun2021semantic, liu2021tripartite}. 
For example, Mask-guided matting~\cite{Yu-2021-MG} takes a general coarse mask as guidance, and proposes a Progressive Refinement Network module to achieve robust guidance.
Matting Anything~\cite{Li-2023-MattingAny} leverages the recent Segment Anything Model (SAM), and further proposes a model that can estimate the alpha matte of any target instance with prompt-based user guidance in an image.

\paragraph{Guidance-free Matting.}
Given the considerable expense associated with acquiring additional guidance, efforts have been made to conduct matting without them, especially for specific foreground scenarios like portraits. These approaches are commonly referred to as guidance-free or trimap-free methods.
Example methods of this type include SHM~\cite{chen2018shm}, SHMC~\cite{liu2020shmc}, HATT~\cite{qiao2020hatt} and GFM~\cite{li2022gfm}.
MODNet~\cite{Ke-2022-MODNet} performs portrait matting by optimizing a series of sub-objectives simultaneously via explicit constraints.
DugMatting~\cite{wu2023dugmatting} explores the explicitly decomposed uncertainties to efficiently and effectively improve matting.
P3M-Net~\cite{MA-2023-P3M-ViTAE} specifically models the interactions between encoders and decoders to perform privacy-preserving portrait matting better.

\paragraph{Diffusion Models.} Our approach builds upon the diffusion model \cite{DDPM, DDIM}, a generative model that has garnered significant attention owing to its exceptional generative capabilities~\cite{rombach2022high}. Diffusion models have also demonstrated remarkable results in text-based image editing tasks, including InstructPix2Pix~\cite{InstructPix2Pix}, Imagic~\cite{kawar2023imagic}, and SINE~\cite{SINE}. In addition, it has been successfully used for various 
tasks~\cite{Xia:2023:DiffIR,Fei:2023:GDP} such as super-resolution~\cite{Yue:2023:ResShift}, inpainting~\cite{Tang:2023:RealFill}, segmentation~\cite{xu2023open,burgert2022peekaboo}. 
Our work represents a pioneering effort in applying diffusion models to image matting.
Compared with the recent concurrent unpublished diffusion-based methods for image matting~\cite{xu2023diffusionmat,hu2023diffusion}, the main difference with our work is the setting. Those methods are trimap-based, while our method facilitates both trimap-free and guidance-based matting. In addition, they are based on the pixel diffusion model, whereas we employ the latent diffusion model (LDM)~\cite{rombach2022high}. The LDM pre-trained on billions of images offers powerful prior. Furthermore, the latent mechanism helps mitigate the impact of potentially imperfect training data, as shown in \figref{teaser} and \figref{imperfect_label}.

\begin{figure*}
    \centering
    \includegraphics[width=1.00\linewidth]{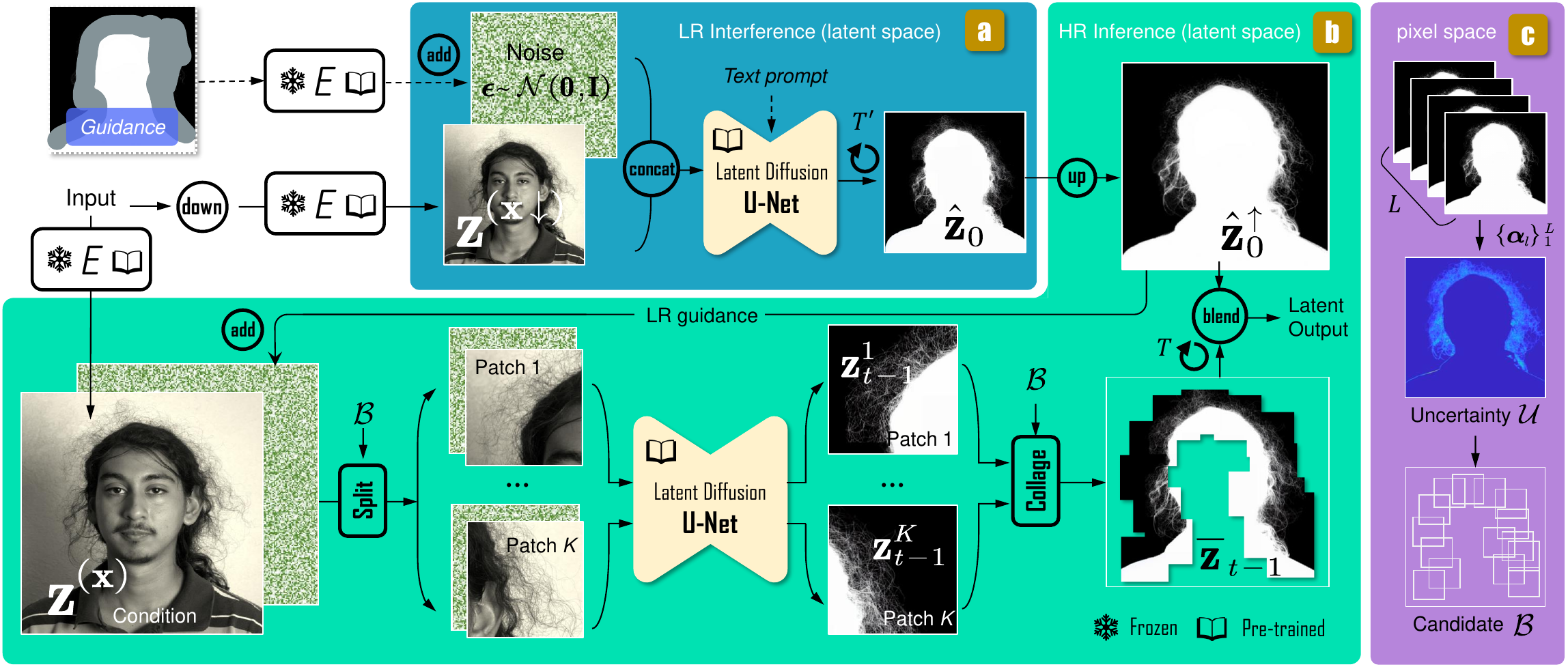}
    \caption{
    Method. (a) The low-resolution inference path can be used alone if we do not need very high-quality mattes or have a limited computational budget. 
    The input is the low-resolution latent feature $\latent^{(\image\downarrow)}$ of the down-sampled image $\image\downarrow$ and
    the sampled noise $\noise_t$.
    If there is spatial guidance $c_\mathcal{S}$ present, we will combine it with the sampled noise as the noisy sample. 
    If a text prompt $c_\mathcal{T}$ is provided, we will deliver it to the U-Net.
    The output of this path is the denoised latent feature $\hat{\latent}_0$. %
    This path requires a few steps $T^\prime\sim10$.
    (c) We run this step multiple times with different random seeds to get $L$ predictions in the pixel space. With them, we estimate the uncertainty map $\mathcal{U}$, and the set of candidate regions $\mathcal{B}=\{b_i\}_1^B$. 
    (b) The high-resolution path. We first add the up-sampled latent feature %
    to the sampled noise. Then, 
    we split the high-resolution latent input and noise into overlapped patches according to $\mathcal{B}$. 
    These patches are respectively fed into the diffusion denoising network. 
    Finally, we merge all denoised patches to get a collage.
    We perform ``split'' and ``collage'' during every denoising step $t\in\{1,\ldots,T\}$.
    We will use a specific text prompt: ``enhance details'' if there is a text prompt used in the LR path. 
    }
    \label{fig:inference}
    \Description{Method.}
\end{figure*}

\section{Method}
\label{sec:method}

We solve the matting problem in a conditional generation manner by training a diffusion model to jointly model the distribution of alpha matte $p(\matte)$ and draw an alpha matte $\matte$ from the distribution conditioned on the input image $\image$. 
Thanks to its generative ability and pre-trained rich image knowledge, our model can find the foreground and generate alpha matte with fine boundary details without guidance (\secref{method_base}).
Our tailored high-resolution inference enables the process of arbitrary-resolution images (\secref{method_hr}).
Besides guidance-free matting, we can seamlessly integrate additional guidance to our trained model, such as a trimap, coarse mask, scribbles, and texts, to alleviate ambiguity in matting (\secref{method_guide}).

\subsection{Generative Formulation}
\label{sec:method_formulation}

We model the distribution of alpha matte $p(\matte)$ with a pre-trained latent diffusion model~\cite{rombach2022high}. 
Given an alpha matte $\matte\sim p(\matte)$, 
we encode it with the pre-trained encoder $\mathcal{E}$ to get its latent representation $\latent^{(\matte)} = \mathcal{E}(\matte)$. 
We then apply the diffusion process to the latent representation.
Let $\latent_0 := \latent^{(\matte)}$, the \emph{forward} process gradually adds a small amount of Gaussian noises to the latent of alpha matte $\latent_0$ in $T$ steps. Therefore, a discrete Markov chain $\{\latent_0, \latent_1, \ldots, \latent_T\}$ is constructed such that
\begin{equation}
\begin{aligned}
\latent_t &= \sqrt{1- \beta_t} \latent^{(\matte)}_{t-1} + \sqrt{\beta_t} \noise_{t-1} = \sqrt{\sigma_t} \latent_0 + \sqrt{1 - \sigma_t} \noise\,,
\end{aligned}
\end{equation}
where the step $t\in\{1,\ldots T\}$, $\noise_{t}, \noise\sim \mathcal{N}(\mathbf{0},\mathbf{I})$ are Gaussian noises and  $\sigma_t := \prod_{s=1}^t \beta_s$. 
The variance schedule $\{\beta_1,\ldots, \beta_T\}$ enables multiple scales of Gaussian noises added to $\latent_0$. 

To model the distribution of $\latent^{(\matte)}$, 
the \emph{backward} process trains a score-based model $\epsilon_\theta$ to predict the noise introduced to the noisy sample $\latent_t$ at step $t$. The objective of training is to minimize
\begin{equation}
    \mathbb{E}_{\epsilon \sim \mathcal{N}(0, 1), t, \latent_0} \big[ \Vert \epsilon_t - \epsilon_\theta(\latent_{t}, t) \Vert_{2}^{2}\big]\,.
\label{eq:loss-1}
\end{equation} 
Training the model on a set of alpha mattes $\{\matte_i\}_{i=1}^{N}\sim p(\matte)$ enables modeling their distribution $p(\matte)$. After training, we can perform ancestral sampling~\cite{song2020score} to generate a sample $\latent_0$ from a normally distributed variable $\latent_T\in\mathcal{N}(\mathbf{0}, \mathbf{1})$. Subsequently, by passing $\latent_0$ through the decoder $\mathcal{D}$, we obtain a matte $\hat{\matte} \sim p(\matte)$.

\subsection{Conditional Generation with a Single Input Image}
\label{sec:method_base}

The matting task aims to produce the alpha matte corresponding to a given input image $\image$, rather than generating a random alpha matte. As a result, we condition the generation process on the input image $\image$.
Specifically, we concatenate the latent $\latent^{(\image)}:=\mathcal{E}(\image)$ of the input image $\image$ with the noisy sample $\latent_t$ and then feed the concatenated tensor to the model. 
To teach the model to generate alpha matte $\matte$ conditioned on the input image $\image$, we train it with paired data $\{(\image_i, \matte_i)\}_{i=1}^{N}\in p_{\text{data}}$ by minimizing:
\begin{equation}
    \mathcal{L} = \mathbb{E}_{\epsilon \sim \mathcal{N}(0, 1), t, \latent_0, \latent^{(\image)}} \big[\Vert \epsilon_t - \epsilon_\theta(\latent_{t}, \latent^{(\image)}, t) \Vert_{2}^{2}\big]\,.
\label{eq:loss-2}
\end{equation}

We initialize model $\epsilon_\theta$ with pre-trained weights from Stable Diffusion (SD)~\cite{rombach2022high}.
The weights learned on billion-level natural images~\cite{schuhmann2022laion} possess extensive knowledge of image semantics and details.
To adapt the denoising score-based model $\epsilon_\theta$ for alpha matte generation, we extend its architecture by duplicating its input layers. The weights of these newly added layers are initialized to 0.
Following this modification, we proceed to fine-tune the denoising score-based model. 
Upon completing the training process, we can draw a sample $\hat{\matte}$ from $p(\matte)$ conditioned on the input image $\image$ with the ancestral sampling and decoding.

\subsection{HR Inference with LR Guidance}
\label{sec:method_hr}
The current image resolution is typically high, often exceeding 2K. Applying the diffusion model to such high-resolution (HR) inputs requires computational resources that are not readily available.
Inference with low-resolution (LR) images is sub-optimal for generating mattes with detailed boundaries.
To address this issue, we propose an HR inference method leveraging patch-based inference.
However, applying patch-based inference, like MultiDiffusion~\cite{bar2023multidiffusion}, to high-resolution matting presents two major challenges: the lack of context and redundant computations. 
To overcome the issue of limited context, we use a predicted low-resolution matte to guide the process. For reducing computational load, we take advantage of the sparsity inherent in alpha mattes.

\paragraph{Patch Sampling}
The diffusion model produces stochastic alpha mattes under different random seeds. 
However, stochastic results generally occur at the boundary regions, where the matte quality is inadequate, while other regions are deterministic and their matte quality is good enough. 
We pay attention to these boundary regions, which represent small portions of the input.
Other regions can be directly determined through up-sampling the matte from LR inference.
Taking advantage of the sparsity of fractional alpha values in the matte, we can reduce computations while maintaining good quality.
First, we downsample the HR image and pass it through the diffusion model, yielding a low-resolution matte prediction $\hat{\matte}_{i}$.
We perform the low-resolution inference $L$ times using varying random seeds, resulting in $L$ low-resolution predictions $\mathcal{A}=\{\hat{\matte}_{1}, \hat{\matte}_{2},\ldots, \hat{\matte}_{L}\}$. Their standard deviation is calculated to approximate the uncertainty map $\mathcal{U}=\sqrt{\mathbb{E}(\mathcal{A}-\mathbb{E}(\mathcal{A}))}$. 
We identify regions on the uncertainty map $\mathcal{U}$ with high information entropy as candidate patches $\mathcal{B}=\{b_i\}_{i=1}^B$ that require refinement. 
As depicted in \figref{inference}, high uncertainty is often observed around complex regions, such as hair boundaries. Thus, based on this uncertainty map, we select candidate regions for further processing.

{
\paragraph{Patch-Based Inference}
We perform inference on the selected patches $\mathcal{B}$.
The noise latent for each patch is not independently sampled.
This strategy ensures consistent prediction for different patches, especially for overlapped areas.
We sample a noise $\latent_T\in \mathcal{N}(\mathbf{0}, \mathbf{1})$ with the same size as the input image.
Then, we crop patches $\{\latent_T^1, \latent_T^2, \ldots, \latent_T^K\}$ from the sampled noise $\latent_T$, where
$\latent_T^k = F(\latent^T | b_i)$ and
$F$ denotes the cropping operator.
The image condition used to condition the model is cropped from the input image's latent similarly.
We feed the noise and image latent of the patch to the diffusion model.
During the ancestral sampling, each step $t$ will produce latent samples $\{\latent_t^1, \latent_t^2, \ldots, \latent_t^K\}$ for patches $\{b_1, b_2, \ldots, b_K\}$. 
Before passing them to the next step $t-1$, we  merge them by
\begin{equation}
    \bar{\latent}_t = \sum_{k=1}^K F^{-1}(\latent_t^k | b_k),
\end{equation}
where $F^{-1}$ is the uncropping operator which puts the latent patch $\latent_t^k$ back to the patch location $b_k$ where it was cropped from the input image.
Then, we get the latent patches for the next step from $\bar{\latent}_t$. After the ancestral sampling, we will obtain the denoised latent $\bar{\latent}_0$.
It is finally merged with the up-sampled LR matte on latent space to get the final alpha matte.
The coarse-to-fine strategy is similar to previous high-resolution matting methods \cite{lin2021real}; however, the main difference is the proposed guidance mechanism for the diffusion model.
}

\paragraph{Guidance Mechanism}
Performing the model on cropped patches often produces flawed results because the model cannot perceive the context information of the whole image and could be misled by local patches. 
To address this issue, we propose to use the predicted LR matte as guidance. 
Although the matte predicted from LR input has imperfect boundary details, it has sufficiently accurate predictions for other regions.
Thus, instead of starting from pure noise $\noise \in\mathcal{N}(\mathbf{0}, \mathbf{1})$, we start the backward process from 
\begin{equation}    
\latent_T=\sqrt{(1- \sigma_T)/\sigma_T}\, \noise + \hat{\latent}_0^{\uparrow},
\end{equation}
where $\hat{\latent}_0^{\uparrow}$ denotes the upsampled latent corresponding to one of the predicted LR mattes $\{\hat{\matte_l}\}_1^L$
This strategy is simple but effective.
During training, $\latent_T$ is a summation of the ground truth alpha matte latent $\latent_0$ and Gaussian noise $\noise$. 
$\latent_0$ contains low-frequency (foreground and background) and high-frequency (boundary) information while $\epsilon_T$ is high-frequency.
The noise will flood the high-frequency information in $\latent_0$. In other words, $\latent_T$ is approximately the combination of the low frequency of $\latent_0$ and the high frequency of $\noise$.
The model learns to extract low-frequency data from noisy samples. 
During inference, the given $\hat{\latent}_0^{\uparrow}$ also contains both low-frequency (foreground and background) and high-frequency (boundary) information. 
The high frequency, which could be inaccurate, is flooded, and the model can extract the correct low frequency. 
This strategy can also facilitate the incorporation of users' guidance 
in the next section.

\subsection{Additional Guidance}
\label{sec:method_guide}

Matting without \emph{any} guidance could lead to ambiguity. 
For example, when there are multiple people in an image, it could be difficult to determine which one to extract.
Additional guidance, such as a human-annotated trimap, a coarse mask derived from semantic segmentation, scribbles, clicks, and a text prompt, would be helpful in this case. 
Our method can incorporate additional guidance if present. 

\paragraph{Text Guidance}
Adding text guidance is relatively easy since we use the text-to-image generative diffusion model.
We annotate the text description of training images with BLIP2~\cite{li2023blip}.
Each annotation describes the target foreground in the training image.
Given the CLIP feature $c_\mathcal{T}$ of the text prompt $\mathcal{T}$, we use the cross-attention mechanism to inject the control into the denoising model. 
We train the denoising model with the annotated prompt by minimizing
\begin{equation}
    \mathcal{L} = \mathbb{E}_{\epsilon \sim \mathcal{N}(0, 1), t, \latent_0, \latent^{(\image)}, c_{\mathcal{T}}} \big[\Vert \epsilon_t - \epsilon_\theta(\latent_{t}, \latent^{(\image)}, c_{\mathcal{T}}, t) \Vert_{2}^{2}\big]\,.
\label{eq:loss-3}
\end{equation}
During training, for small patches, we use a specific prompt ``enhance details'' instead of avoiding confusing the model.

\paragraph{Spatial Guidance}
Spatial guidance like a trimap, coarse mask~\cite{Yu-2021-MG}, and scribbles are more popular than text prompts for image matting.
Inspired by the guidance mechanism described in \secref{method_hr}, we use a similar method for injecting spatial guidance $\mathcal{S}$. 
We extract this guidance's latent representation $c_{\mathcal{S}}$ and a mask indicating unknown regions $m_\text{unknown}$.
For coarse mask, $m_\text{unknown} = \textbf{I}$ and for scribble $m_\text{unknown}$ represents regions without scribbles.
At inference, we perform ancestral sampling from 
\begin{equation}    
\latent_T= \sqrt{(1- \sigma_T)/\sigma_T}\, \noise + (1 - m_\text{unknown})\, c_{\mathcal{S}}\,.
\end{equation}
We can apply various kinds of guidance\footnote{The domain of the guide image should match that of the alpha matte.} directly at the inference time without training with them.

\section{Experiments}
\label{sec:exp}

\subsection{Protocol}

\paragraph{Dataset.}
We conduct experiments on real-world datasets rather than synthetic ones.
We train our model on the training set of {P3M-10K~\cite{Li-2021-P3M}}, a dataset containing 9,421 high-resolution real-world face-blurred portrait images and human-annotated alpha mattes that are not perfectly accurate. 
We evaluate the performance on three benchmarks:
{P3M-P} dataset containing 500 face blurred images. 
Each image has a corresponding trimap and alpha matte.
They are validation sets from {P3M-10K} that share a similar distribution of alpha matte with the training set.
{PPM-100~\cite{Ke-2022-MODNet}} is a dataset with 100 high-resolution images with corresponding fine annotations.
{RVP~\cite{Yu-2021-MG}} consists of 636 portraits with alpha mattes and coarse segmentation masks.

\paragraph{Compared methods.}
We compare our approach against several state-of-the-art matting methods.
\textbf{Guidance-free:}
MODNet~\cite{Ke-2022-MODNet}, P3M~\cite{Li-2021-P3M}, ViTAE-S~\cite{MA-2023-P3M-ViTAE}. 
These methods, except for MODNet\footnote{
We utilize the publicly available checkpoints released by MODNet. 
It was trained on proprietary datasets. 
We attempted to align MODNet’s training data with ours. 
However, this reproduction resulted in unsatisfactory outcomes. 
}, are trained on P3M-10K. 
\textbf{Trimap-based:} IndexNet~\cite{lu2019indices}, MatteFormer~\cite{park2022matteformer}, and DiffMat~\cite{xu2023diffusionmat}, which is a concurrent %
method using diffusion models for trimap-based matting.
\textbf{Mask Guided Matting:} MAM~\cite{Li-2023-MattingAny}, which incorporates SAM \cite{kirillov2023segany} as its backbone, and MG-Mat~\cite{Yu-2021-MG}.

\paragraph{Metrics.}
Evaluation metrics include the Sum of Absolute Differences (SAD), Mean Squared Error (MSE), Mean Absolute Difference (MAD), 
and Connectivity (Conn.)~\cite{rhemann2009perceptually}. 
We apply all metrics on whole images.
We scale the MAD and MSE values by a factor of 10\textsuperscript{3}.

\begin{table}[t]
    \centering
    \caption{{Quantitative results of trimap-free portrait matting.} We compare our method with trimap-free portrait matting methods. $^\dagger$For the trimap-based method DiffMat, we provide a mask with all pixels labeled unknown.
    $^*$We removed ambiguous samples from the dataset, 5 out of 100 from PPM and 20 out of 500 from P3M-P, which will be elaborated on in the supplementary. 
    }
    \label{tab:benchmark-tab}
    \setlength\tabcolsep{2pt}
    \resizebox{\linewidth}{!}{
    \begin{tabular}{lrrrrrrrrrrrr}
        \toprule
      & \multicolumn{4}{c}{{PPM$^*$}} & \multicolumn{4}{c}{{P3M-P$^*$}} \\\cmidrule(lr){2-5}\cmidrule(lr){6-9}
      & \footnotesize MSE $\downarrow$ & \footnotesize MAD $\downarrow$ & \footnotesize SAD $\downarrow$ & \footnotesize Conn $\downarrow$ & \footnotesize MSE $\downarrow$ & \footnotesize MAD $\downarrow$ & \footnotesize SAD $\downarrow$ & \footnotesize Conn $\downarrow$\\\midrule
      \small DiffMat$^\dagger$~\cite{xu2023diffusionmat} & \footnotesize522.1 & \footnotesize594.9 & \footnotesize5681.3 & \footnotesize5623.9 & \footnotesize510.2 & \footnotesize582.7 & \footnotesize999.1 & \footnotesize989.0 \\
      \small MODNet~\cite{Ke-2022-MODNet}   & 4.5 & 10.1 & 96.0 & 81.1 &  11.3 &
17.4 & 29.9& 26.6 \\
      \small P3M~\cite{Li-2021-P3M} & 5.8 & 9.6 & 93.3 & 96.1 &  2.7 &  5.1 &  8.8 & 8.3 \\
      \small ViTAE-S~\cite{MA-2023-P3M-ViTAE} & \underline{3.4} & \underline{6.5} & \underline{62.6} & \underline{59.3} & \underline{1.8} & \underline{4.3} & \underline{7.4} & \underline{7.2} \\
      \small Ours & \textbf{2.5} & \textbf{6.3} & 
     \textbf{56.9} & \textbf{54.0} & \textbf{1.6} & \textbf{4.1} & \textbf{7.1} & \textbf{6.8} \\
      \bottomrule
    \end{tabular}
    }
\end{table}

\begin{table}[t]
    \centering
    \caption{{Quantitative results of guidance-based portrait matting.} 
    We compare our method with guidance-based portrait matting methods. 
    The guidance is a mask in the top portion of the table, while in the bottom portion, it is a trimap.
    $^\dagger$P3M-P does not provide the segmentation mask; therefore, we use coarse masks extracted from trimaps $m$ as guidance ($m[m>=0.5] = 1$). Although our scores are worse than some methods since the label is imperfect, our visual results are better. Note that P3M has the same distribution as the training set; therefore, it could not reflect overfitting the imperfect labels.
    }
    \label{tab:guidance-based}
    \setlength\tabcolsep{2pt}
    \resizebox{\linewidth}{!}{
    \begin{tabular}{lrrrrrrrrrrrr}
        \toprule
      & \multicolumn{4}{c}{{RVP}} & \multicolumn{4}{c}{{P3M-P$^\dagger$}} \\\cmidrule(lr){2-5}\cmidrule(lr){6-9}
      & \footnotesize MSE $\downarrow$ & \footnotesize MAD $\downarrow$ & \footnotesize SAD $\downarrow$ & \footnotesize Conn $\downarrow$ & \footnotesize MSE $\downarrow$ & \footnotesize MAD $\downarrow$ & \footnotesize SAD $\downarrow$ & \footnotesize Conn $\downarrow$\\\midrule
      \small MAM~\cite{Li-2023-MattingAny} & 20.7 & 36.3 & 48.5& 44.6 & 7.9 & 13.4 & 23.1 & 20.5\\
      \small MG-Mat~\cite{Yu-2021-MG} & \textbf{9.4} & \underline{20.7} & \underline{29.2} & \textbf{25.5} & 5.7 & 12.8 & 22.0 & 18.4\\
      \small DiffMat~\cite{xu2023diffusionmat} & 16.6 & 32.5 & 44.4 & 41.2 & 45.0 & 49.5 & 84.4 & 84.5\\
      \small Ours \small - mask  & \underline{11.6} & \textbf{19.6} & \textbf{26.7} & \underline{26.1} & {1.6} & {4.2} & {7.2} & {6.2} \\\midrule
      \small IndexNet~\cite{lu2019indices} & -- & -- & -- & -- & \underline{1.2} & 4.2 & 7.0 & \underline{6.0} \\ %
      \small MatteFormer~\cite{park2022matteformer} & -- & -- & -- & -- & 1.4 & 4.1 & 7.1 & 6.5 \\ %
      \small DiffMat~\cite{xu2023diffusionmat} & -- & -- & -- & -- & \textbf{1.0} & \textbf{3.6} & \textbf{6.2} & \textbf{5.2}\\
      \small Ours \small - trimap & -- & -- & -- & -- & {1.6} & \underline{4.0} & \underline{6.9} & \underline{6.0}\\
      \bottomrule
    \end{tabular}
    }
\end{table}

\newlength{\qualfw}
\setlength{\qualfw}{0.25\linewidth}

 \newcommand{\qualfrow}[3]{
        \resizebox{0.33\textwidth}{!}{%
        \begin{tabular}{@{}c@{\,\,}c@{}}
        & 
        \includegraphics[width=1.365\qualfw]{figure/zoom_results/#1_#2_patch_box.jpg} 
        \end{tabular}
        \begin{tabular}{@{}c@{\,\,}c@{c}}
        
        \includegraphics[width=\qualfw]{figure/zoom_results/#1_#2_patch.jpg}\\
    \includegraphics[width=\qualfw]{figure/zoom_results/#1_#2_patch_2.jpg}\\ 
    
    \end{tabular}
}
}
 \newcommand{\qualfrowcom}[3]{
        \resizebox{0.33\textwidth}{!}{%
        \begin{tabular}{@{}c@{\,\,}c@{}}
        & 
        \includegraphics[width=1.365\qualfw]{figure/composition/#1_comp_#2.jpg} 
        \end{tabular}
        \begin{tabular}{@{}c@{\,\,}c@{c}}
        
        \includegraphics[width=\qualfw]{figure/zoom_results/#1_#2_patch.jpg}\\
    \includegraphics[width=\qualfw]{figure/zoom_results/#1_#2_patch_2.jpg}\\ 
    
    \end{tabular}
}
}

\begin{figure*}
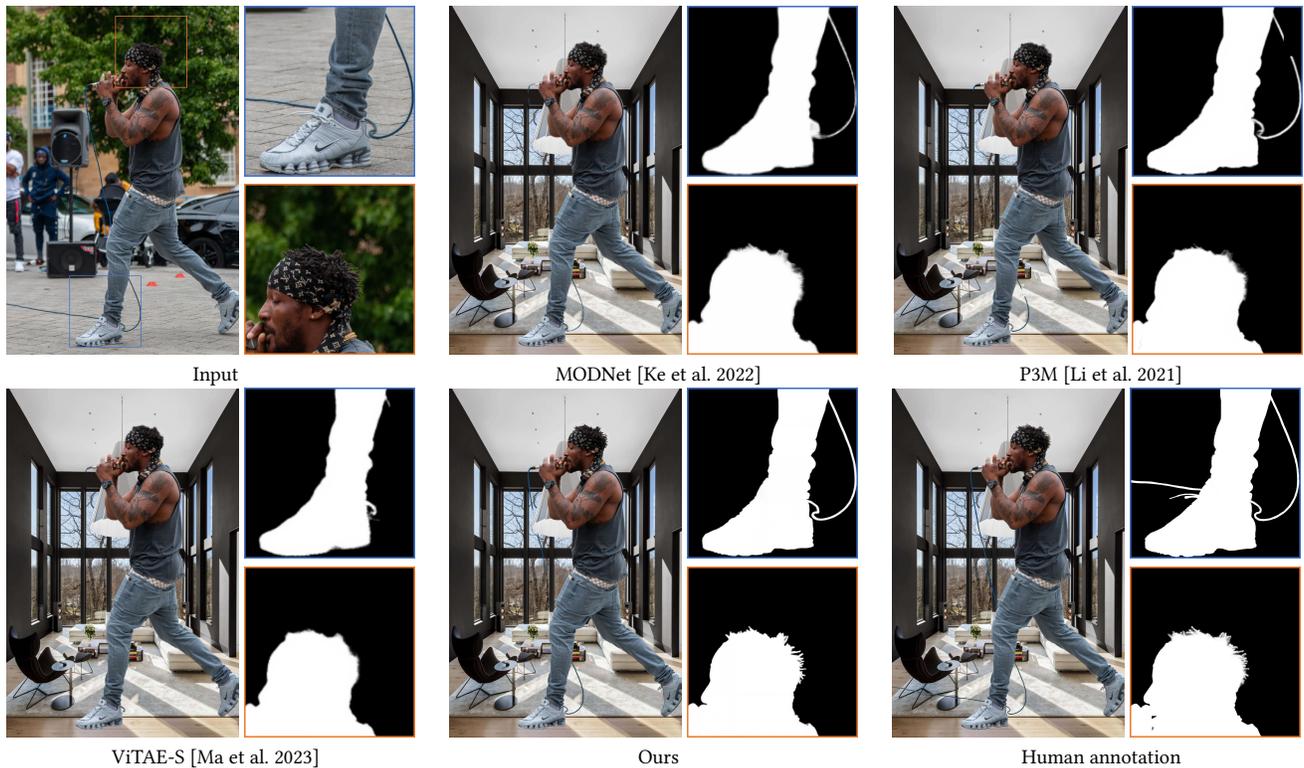

    \centering
    \small
    \resizebox{1.00\textwidth}{!}{
     \begin{tabular}
    {
        @{\hspace{0mm}}c@{\hspace{1.2mm}} 
        @{\hspace{0mm}}c@{\hspace{1.2mm}}
        @{\hspace{0mm}}c@{\hspace{1.2mm}}
        @{\hspace{0mm}}c@{\hspace{1.2mm}}
        @{\hspace{0mm}}c@{\hspace{1.2mm}} 
        @{\hspace{0mm}}c@{\hspace{1.2mm}} 
    }

    \qualfrow{50238484256_6c8d245d7a_o}{img}{}
    & \qualfrowcom{50238484256_6c8d245d7a_o}{modnet}{}
    & \qualfrowcom{50238484256_6c8d245d7a_o}{p3m}{}\\
    Input & MODNet~\cite{Ke-2022-MODNet} & P3M~\cite{Li-2021-P3M} \\
    \qualfrowcom{50238484256_6c8d245d7a_o}{p3m_net}{} &   \qualfrowcom{50238484256_6c8d245d7a_o}{ours}{} & 
    \qualfrowcom{50238484256_6c8d245d7a_o}{gt}{} \\
    ViTAE-S~\cite{MA-2023-P3M-ViTAE} & Ours & Human annotation\\
    \\

    \end{tabular}
    }
    \caption{Visual results of trimap-free matting on PPM-100~\cite{Ke-2022-MODNet}. Our method achieves more accurate matting results, especially around thin and detailed structures, compared to prior work. We extracted the foreground using the technique proposed by Germer~\etal~\cite{germer2021fast} and composited it onto a new background sampled from a public background database \cite{lin2021real}.}
    \label{fig:visual1}
    \Description{Visual results of trimap-free matting on PPM-100}
\end{figure*}

\subsection{Trimap-free Matting}

\tabref{benchmark-tab} presents the quantitative results for trimap-free setting. Our method achieves the best scores in all metrics. 
It consistently delivers good results on three benchmarks,
showcasing its robustness and versatility across diverse scenarios. 
Our approach outperforms established methods like MODNet, P3M, ViTAE-S, and MAM, particularly regarding accuracy and boundary detail handling. This is evident in the lower SAD, MSE, MAD, and improved Connectivity scores compared to the competing methods. These results highlight the effectiveness of our generative modeling approach and the use of pre-trained diffusion models in addressing the complexities of trimap-free matting, especially in challenging cases involving intricate details and varying image qualities.

\figref{visual1} presents the qualitative comparisons.
The input shows a complex scene with a person that includes intricate details like hair and shoelaces, which are challenging for matting algorithms. As highlighted by the insets, MODNet and P3M outputs lack fine detail, particularly in the hair and feet regions. In contrast, the results from ViTAE-S, while quantitatively close to our method, visually lack the nuanced details that our approach captures. Our result is more similar to the ground truth, including a clear and precise boundary matte, which faithfully reproduces the fine details of the subject, such as individual hair strands and the shoe's silhouette. 
This is evident even in cases where quantitative scores are similar, showcasing the added benefit our approach brings in creating high-fidelity human boundary mattes.

\figref{visual2} showcases the visual results
on the RVP dataset, focusing on a challenging scenario involving complex hair details against a sunset backdrop. The input image presents significant matting difficulty due to the intricate hair strands silhouetted against the varying tones of the sky. MODNet and P3M results display notable artifacts and fail to capture the finer hair details, as evidenced in the zoomed insets. ViTAE-S, although quantitatively competitive, visually lacks fidelity in reproducing the hair's fine structure, as the comparison with the ground truth reveals a less accurate matte. Our method, on the other hand, shows a remarkable capture of detail, closely mirroring the ground truth. The insets highlight our approach's capability to preserve the delicate strands of hair and the subtle nuances in the silhouette, which are critical for a realistic matting outcome. Despite their close quantitative scores, this visual comparison underscores the qualitative edge of our method over ViTAE-S, illustrating our approach's advanced ability to generate detailed human boundary mattes that are distinct and more aligned with the actual scene.

\setlength{\qualw}{0.23\columnwidth}
\newcommand{\qualrowz}[5]{            
	{\includegraphics[trim={#2 #3 #4 #5}, clip,width=\qualw]{prompting/#1.jpg}} &
        {\includegraphics[trim={#2 #3 #4 #5}, clip,width=\qualw]{prompting/#1_no_prompt.jpg}} &
         {{\scriptsize``foreground person\ldots''}}
        &
        {\includegraphics[trim={#2 #3 #4 #5}, clip,width=\qualw]{prompting/#1_full_prompt.jpg}}
}

\newcommand{\qualrow}[5]{            
        {\includegraphics[trim={#2 #3 #4 #5}, clip,width=\qualw]{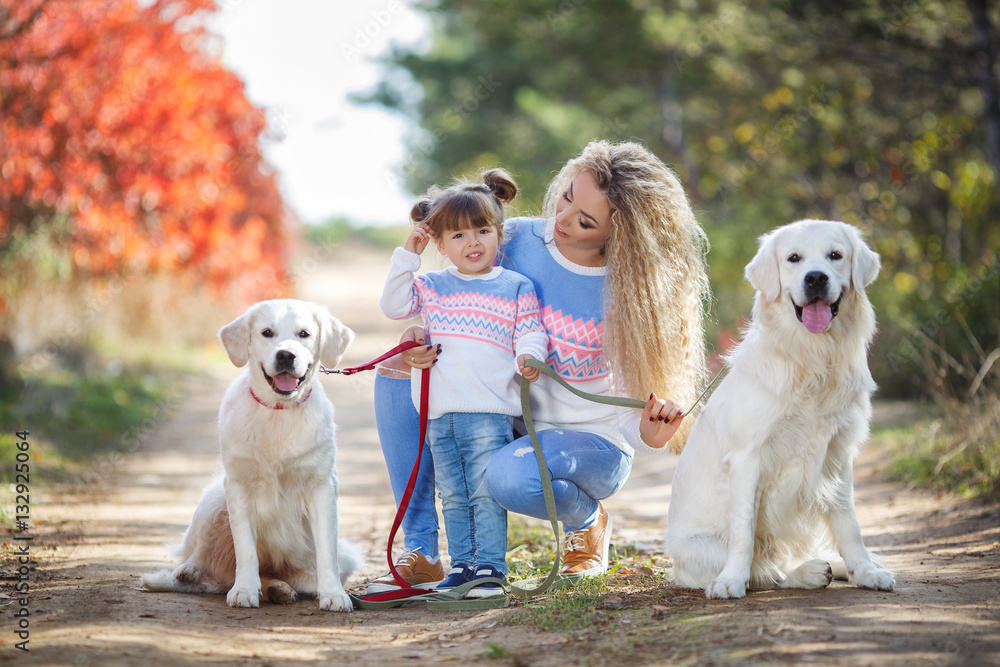}} &  
        {\includegraphics[trim={#2 #3 #4 #5}, clip,width=\qualw]{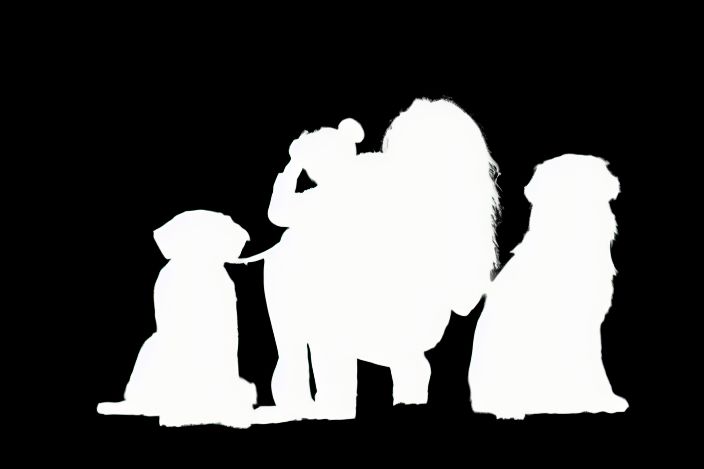}} &         
        {\includegraphics[trim={#2 #3 #4 #5}, clip,width=\qualw]{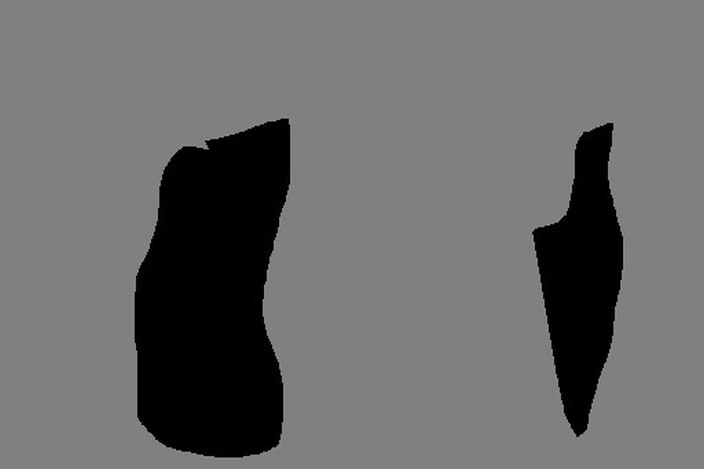}} &
        {\includegraphics[trim={#2 #3 #4 #5}, clip,width=\qualw]{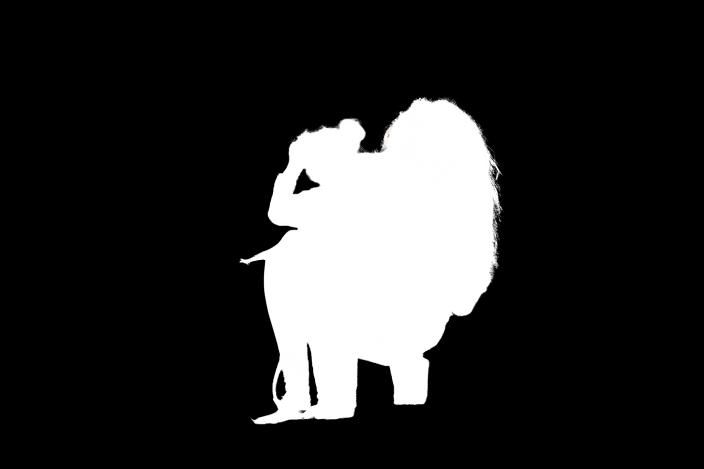}}\\
}
\begin{figure}
    \small
    \centering
    \begin{tabular}
    {   
        @{\hspace{0mm}}c@{\hspace{1.2mm}} 
        @{\hspace{0mm}}c@{\hspace{1.2mm}}
        @{\hspace{0mm}}c@{\hspace{1.2mm}}
        @{\hspace{0mm}}c@{\hspace{1.2mm}}
        @{\hspace{0mm}}c@{\hspace{1.2mm}} 
    }
         \qualrowz{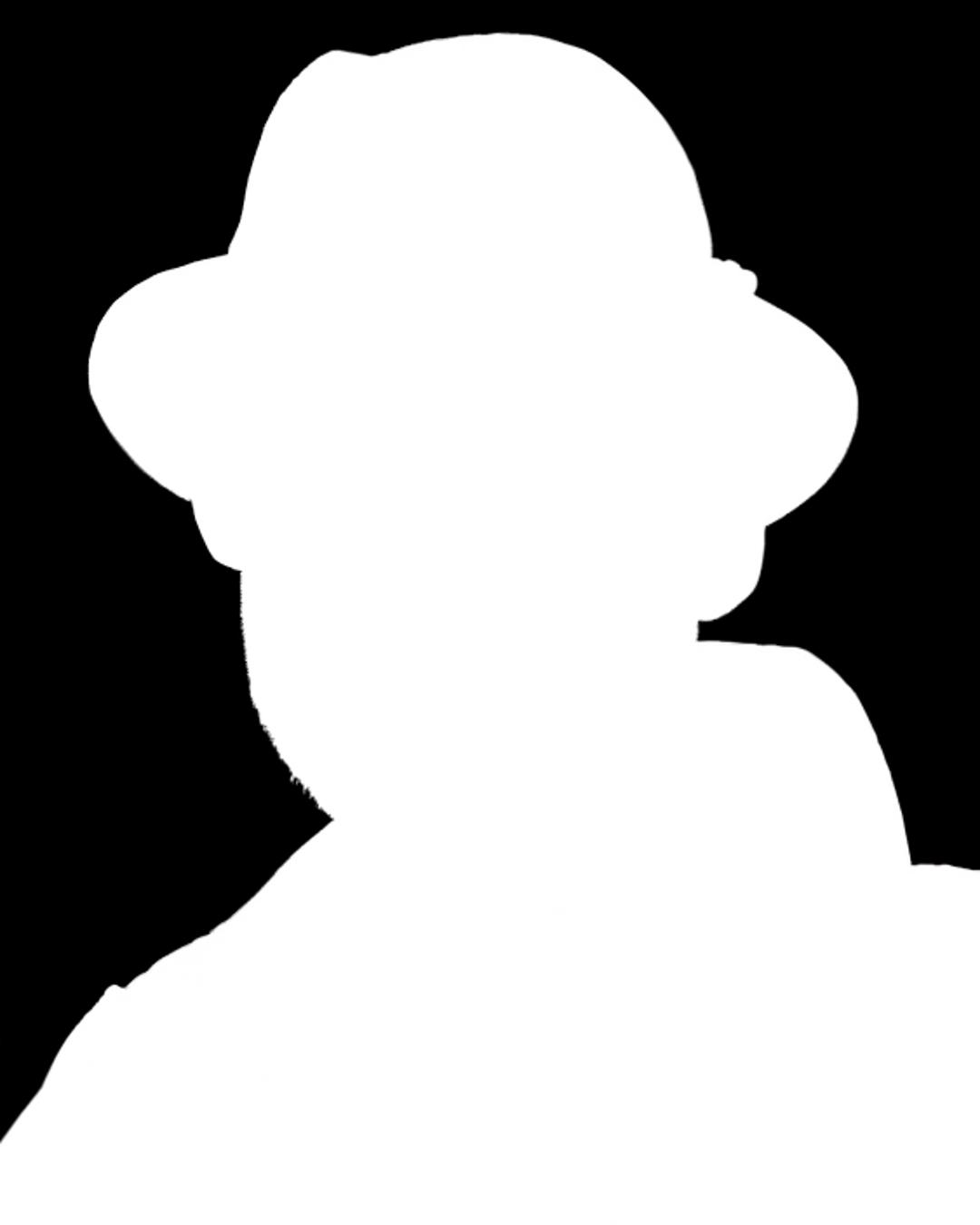}{0}{0}{0}{0}\\
         \qualrow{21}{0}{0}{0}{0}
         Input & w/o guidance & Guidance & w/ guidance \\
         
    \end{tabular}
    \caption{Use of guidance. With various guidance, we can reduce ambiguity.}
    \Description{Use of guidance.}
    \label{fig:promopt}
\end{figure}

\subsection{Guidance-based Matting}

When the foreground is ambiguous, it is inherently challenging for trimap-free matting. In contrast, our method can also use guidance such as text prompt and coarse mask to reduce ambiguity (\figref{promopt}).

\tabref{guidance-based} shows the quantitative comparisons for the guidance-based setting.  
Although the trimap-based method---DiffMat~\cite{xu2023diffusionmat}---performs better than our method in the trimap-based setting, it relies on the high quality of the trimaps. They would fail when using coarse guidance, such as masks from semantic segmentation.
The mask-based method MAM~\cite{Li-2023-MattingAny} refines the mask generated from SAM~\cite{SAM}. Benefiting from SAM, it can predict alpha mattes for ``any'' foreground objects. However, its matting performance is sub-optimal.

\figref{guidancematting} shows the qualitative comparison under the trimap-based setting.
The scores of DiffMat~\cite{xu2023diffusionmat} are better, but we notice their visual results are worse than ours. 
At the same time, the imperfection of the ground-truth mattes is also observed.
We suppose the worse results of DiffMat compared to us are because they are based on the pixel diffusion model, which overfits the training labels.

\section{Analyses and Discussions}

\begin{table}[]
    \centering
    \caption{Ablation study. We implement four variants of our method and conduct the ablation study on PPM-100: (1) Our model without using the pre-trained SD weights; (2) Training with the same prompt for all cropped patches from an image; (3) Our model trained with resized full image rather than patches with different scales; (4) Adding pixel losses to our training phase.
    }
    \label{tab:ablation}
    \small
    \begin{tabular}{lrr}
        \toprule
         & MSE & MAD \\\midrule
        Ours & \textbf{2.5} & \textbf{6.3} \\
        - (1) \emph{w/o} denoising prior & 63.6 & 71.1\\
        - (2) \emph{w/o} specific patch prompt & 38.6 & 46.8\\
        - (3) \emph{w/o} multi-scale data & 8.6 & 15.1\\
        + (4) pixel loss & 58.0 & 65.1\\
        \bottomrule
    \end{tabular}
\end{table}

\paragraph{Ablation studies.}
\tabref{ablation} demonstrates the importance of the pre-trained generative prior, the prompting and the multi-scale training strategy (will be elaborated on in the supplementary), and operating in latent space.
Stable Diffusion (SD) with vast pre-trained knowledge significantly induces semantic information. \tabref{ablation} indicates, without this, training the diffusion model is prone to lose semantic information, \eg resulting in an incomplete person.
Besides semantic information, this generative prior enables us to hallucinate details, \eg, hair boundary. Without it, we could converge to the existing methods. 
Besides, operating within the latent space adeptly preserves essential details, crucial for producing high-quality mattes with emphasis on boundary regions.
This is an advantage of our method over concurrent works built on pixel-based diffusion models.
\figref{full-res} shows the effectiveness of HR inference with LR guidance.

\begin{figure}
    \centering
    \includegraphics[width=0.8\columnwidth]{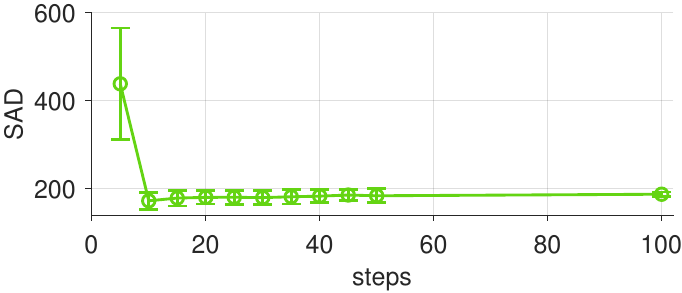}
    \caption{Randomness test. We use 5 different random seeds to test the model on selected images. With the increase of diffusion steps, the mean and std. of SAD error decrease.}
    \label{fig:randomness}
    \Description{Randomness test.}
\end{figure}

\paragraph{Effect of randomness.}
\figref{randomness} depicts that with the increases of steps, the randomness decreases. Besides, we notice that infer with larger patches will also reduce the randomness.

\paragraph{Soft matte.}
Our model can produce soft mattes for out-of-focus blur, as shown in Figure~\ref{fig:blur}, even though the training dataset does not contain annotations for such blur.

\setlength{\qualw}{0.24\columnwidth}
\renewcommand{\qualrow}[5]{     
{\includegraphics[trim={#2 #3 #4 #5}, clip,width=\qualw]{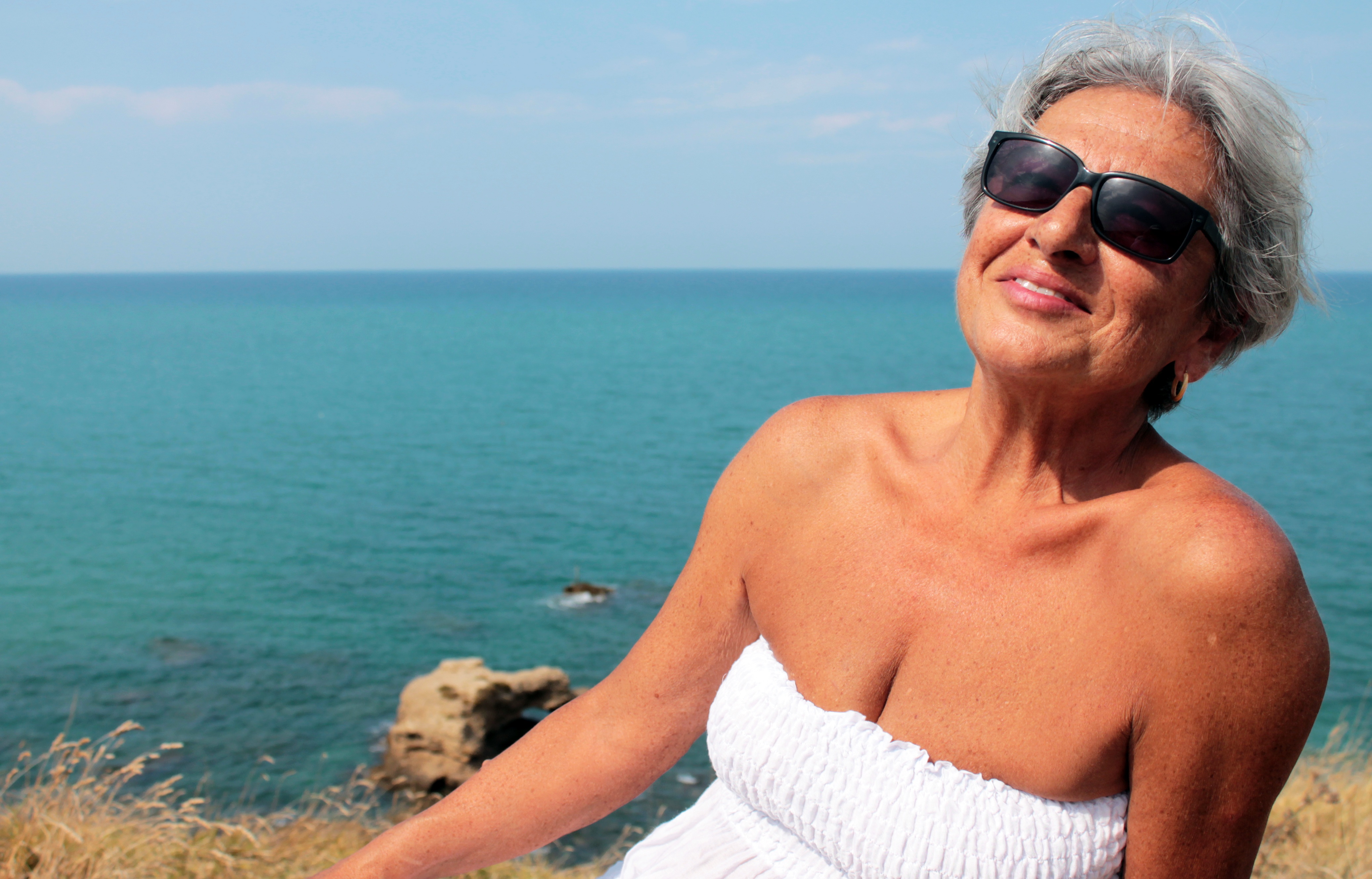}} &
	{\includegraphics[trim={#2 #3 #4 #5}, clip,width=\qualw]{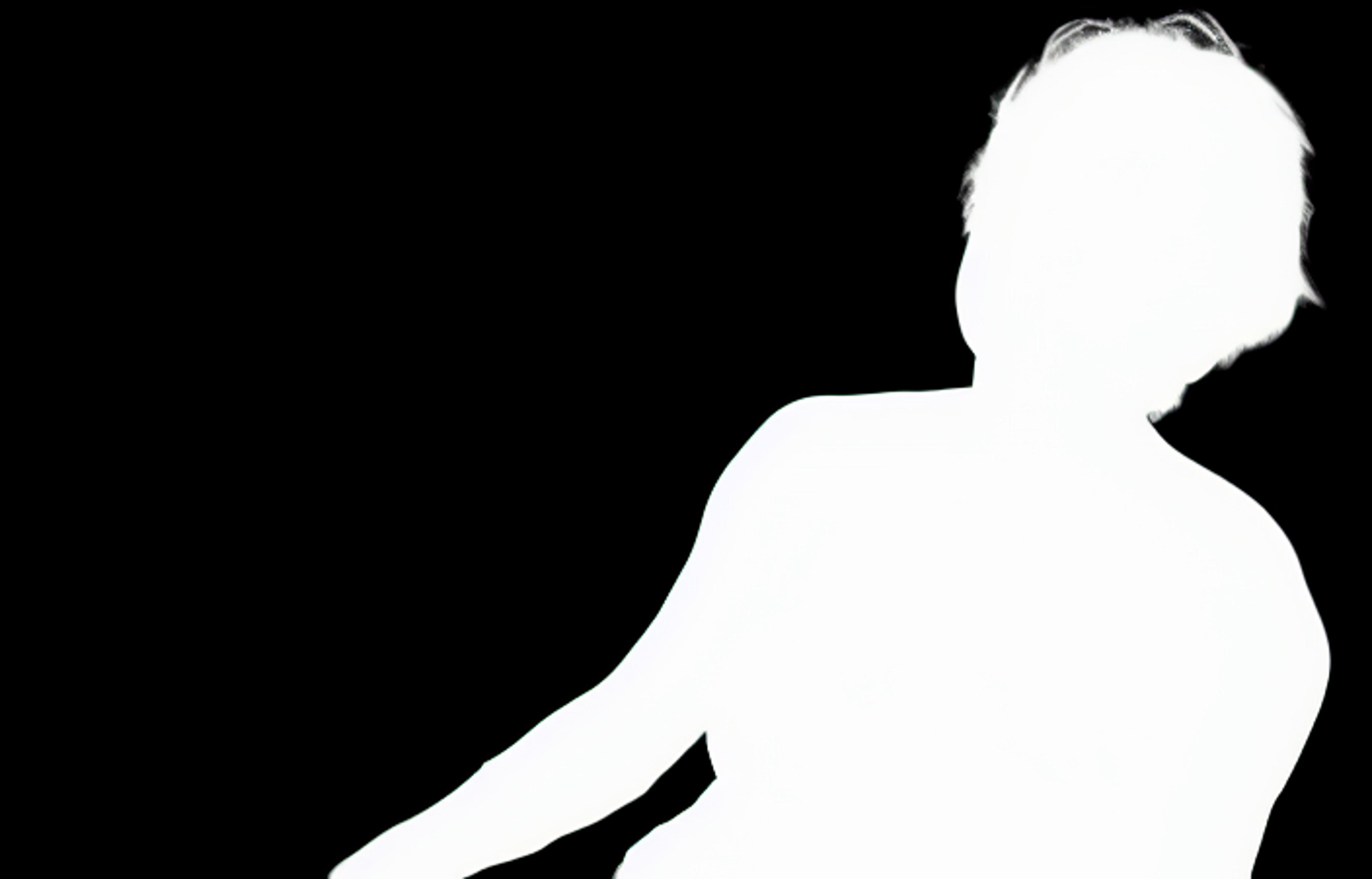}} &
 {\includegraphics[trim={#2 #3 #4 #5}, clip,width=\qualw]{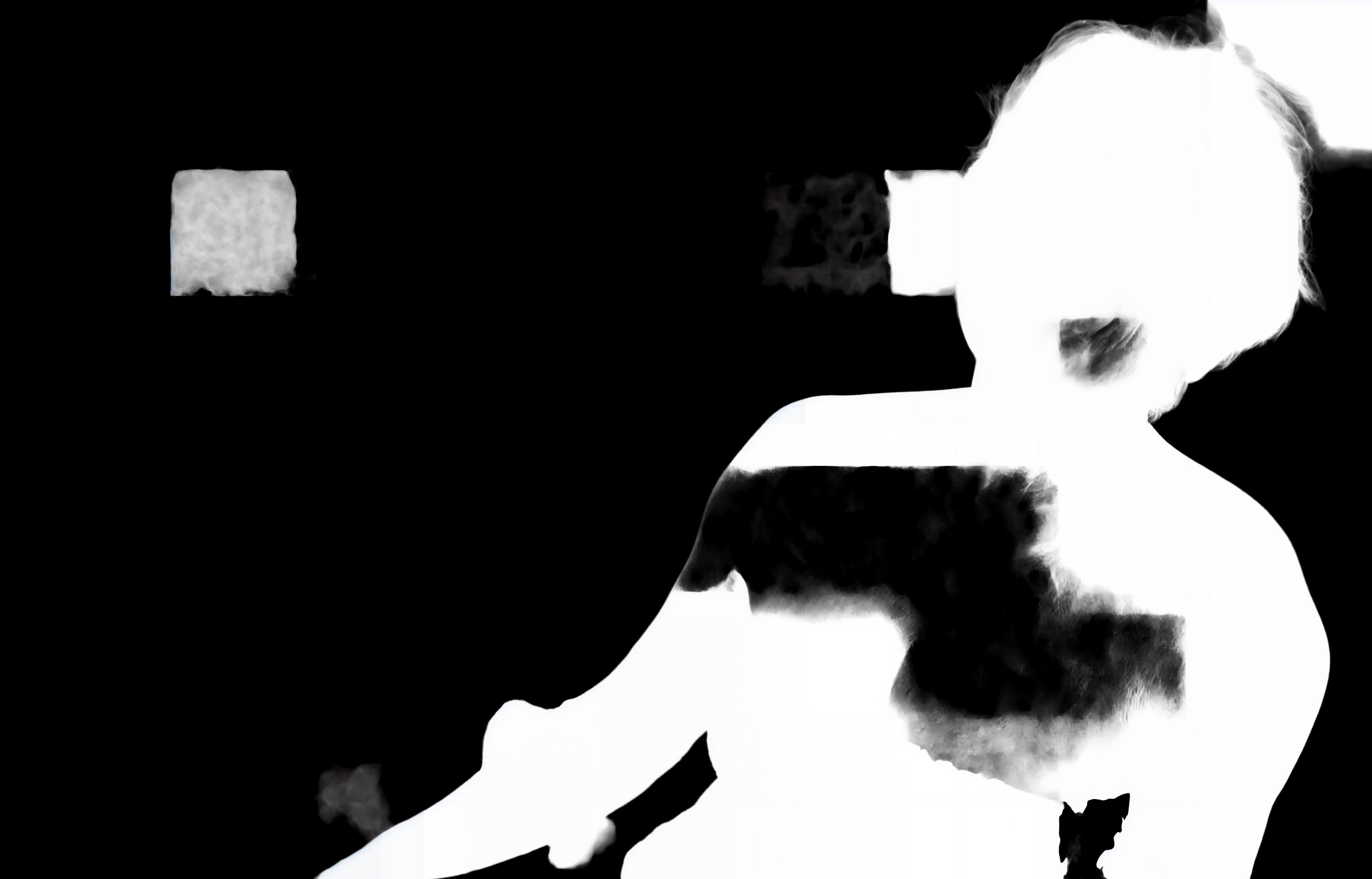}} &
 {\includegraphics[trim={#2 #3 #4 #5}, clip,width=\qualw]{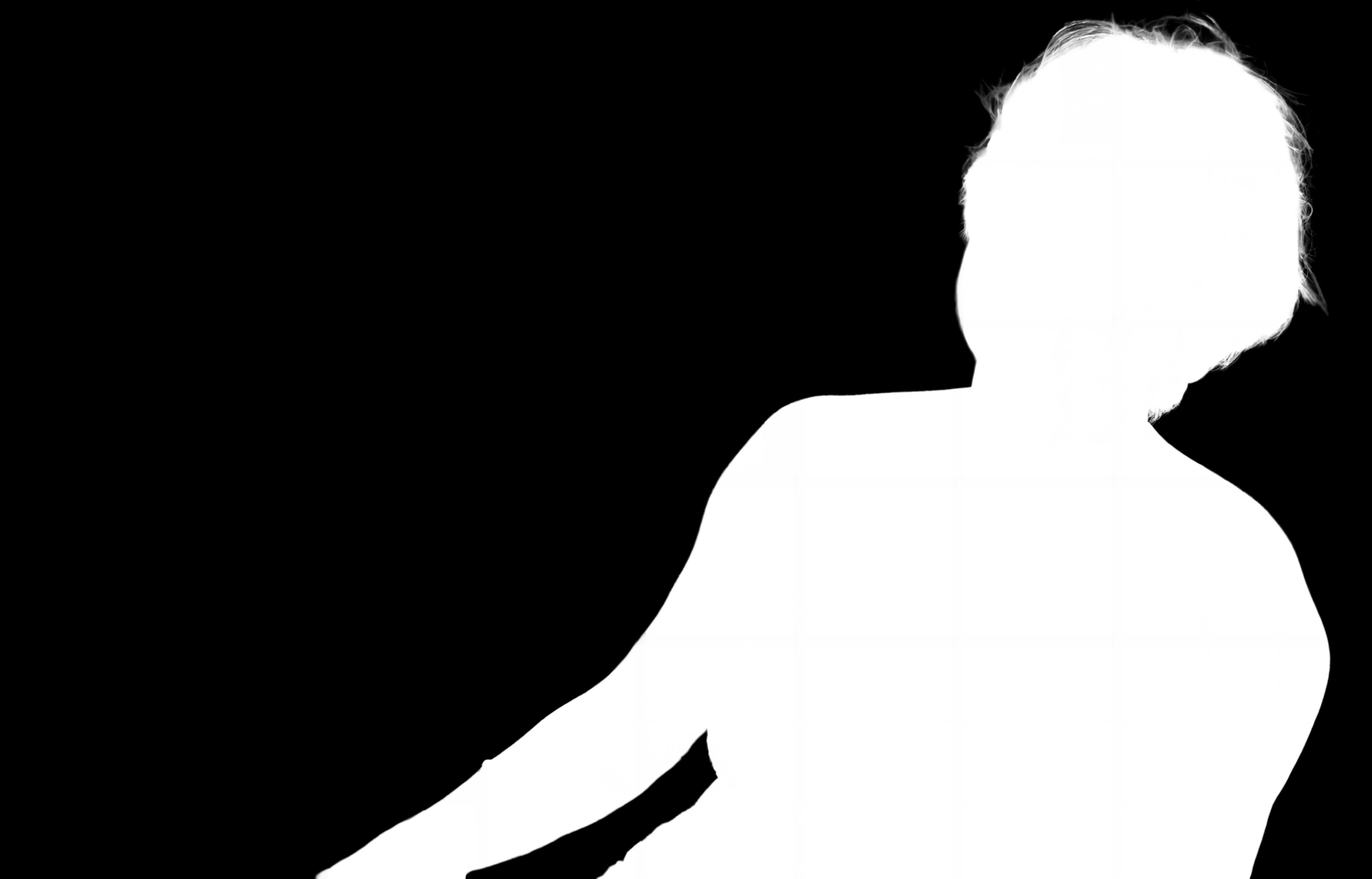}} \\
 Input & LR & w/o guidance & w/ guidance\\
 
}
\begin{figure}
    \centering
    \small
     \begin{tabular}
    {   
        @{\hspace{0mm}}c@{\hspace{1.2mm}} 
        @{\hspace{0mm}}c@{\hspace{1.2mm}}
        @{\hspace{0mm}}c@{\hspace{1.2mm}}
        @{\hspace{0mm}}c@{\hspace{1.2mm}}
        @{\hspace{0mm}}c@{\hspace{1.2mm}}
    }
    \qualrow{}{0cm}{0cm}{0cm}{0cm}\\
    \end{tabular}
    \caption{HR inference with LR guidance.}
    \label{fig:full-res}
    \Description{HR inference with LR guidance.}
\end{figure}

\setlength{\qualw}{0.34\columnwidth}
\renewcommand{\qualrow}[5]{            
	{\includegraphics[trim={#2 #3 #4 #5}, clip,width=\qualw]{visual/#1_lr.jpg}} 
 &
 {\includegraphics[trim={#2 #3 #4 #5}, clip,width=\qualw]{visual/#1_modnet.jpg}} &
 {\includegraphics[trim={#2 #3 #4 #5}, clip,width=\qualw]{visual/#1_p3m.jpg}} & 
 
        {\includegraphics[trim={#2 #3 #4 #5}, clip,width=\qualw]{visual/#1_p3m-net.jpg}} &
        {\includegraphics[trim={#2 #3 #4 #5}, clip,width=\qualw]{visual/#1_ours.jpg}} &
        {\includegraphics[trim={#2 #3 #4 #5}, clip,width=\qualw]{visual/#1_gt.jpg}}\\
}

\begin{figure}
    \centering
    \includegraphics[width=1.00\linewidth]{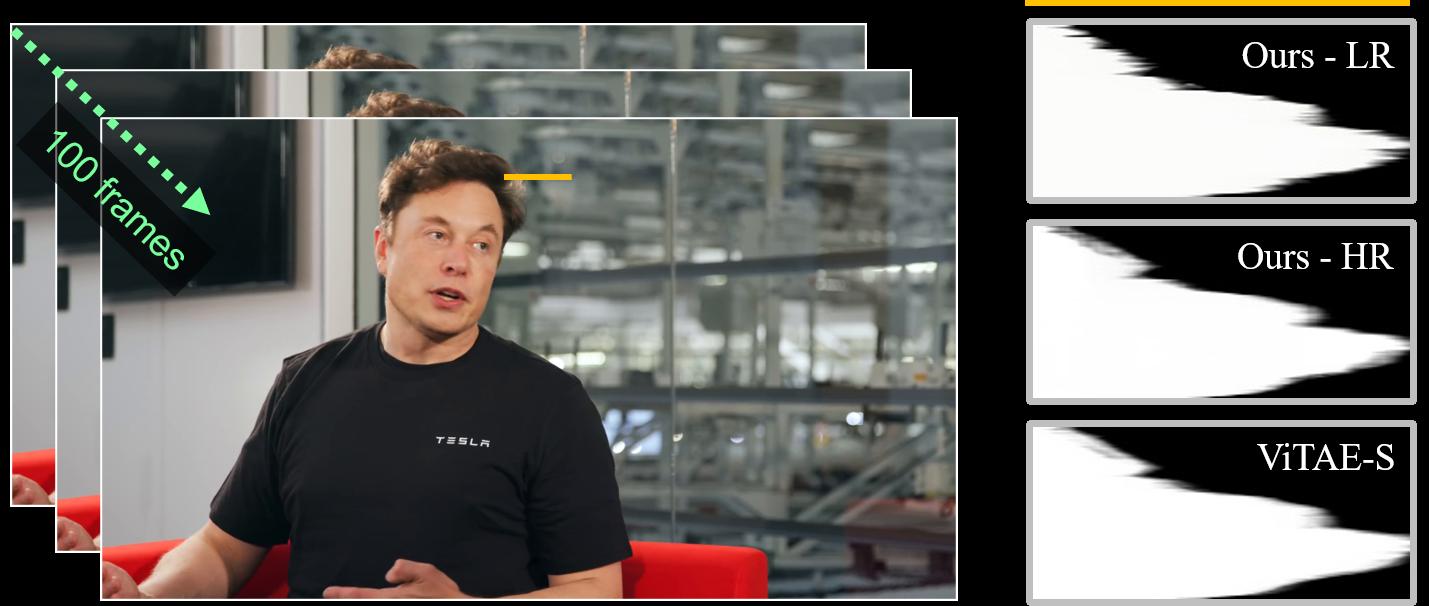}
    \caption{Video inference. 
    By individually processing downsampled frames, our method produces temporal inconsistency in videos. While employing high-resolution frames mitigates this issue, it still suffers from problems similar to regression-based methods.}
    \label{fig:temporal_analysis}
    \Description{Video inference.}
\end{figure}

\paragraph{Limitations.}
Firstly, while our method reduces inference time for HR images compared to naive approaches, it is important to clarify that the inherent limitations of the diffusion model make it less efficient than prior regression-based methods.
Processing a 512$\times$512 images with 50 steps requires about 5s on a NVIDIA V100 GPU card.
However, it is worth noting that some ongoing research efforts are focused on enhancing sampling efficiency, which could help mitigate this limitation \cite{li2023snapfusion,song2023consistency}.
Secondly, our model trained on portrait datasets shows potential for adaptation to other domains, such as animal matting (see Figure~\ref{fig:beyound}), but is unsuitable for matting scenarios with markedly different characteristics, such as subjects like fire. 
Thirdly, our method is designed for image matting and cannot guarantee temporal consistency for videos (see Figure~\ref{fig:temporal_analysis}). Enhancing temporal coherence remains a subject for future research.

\section{Conclusions}
\label{sec:conc}

Our approach presents a straightforward yet highly efficient technique for matting. 
It can perform both trimap-free and guidance-based image matting tasks.
By reframing the problem as a generative task and leveraging diffusion models enriched with pre-trained knowledge for regularization, we have devised innovative designs that empower our model to produce high-resolution and high-quality results. 
Our experimental results on three benchmark datasets not only demonstrate the efficacy of our method in quantitative terms but also showcase its exceptional visual performance, making it a promising solution for the field of matting.

\begin{acks}
We thank anonymous reviewers for their constructive feedback. We also thank Yutong Dai and Zhanghan Ke for their help. This research was supported by NTU112L9009, NTU113L894003, NSTC112-2634-F-002-006, MOST110-2221-E-002-124-MY3, JSPS KAKENHI Grant Number JP23K24876, JST ASPIRE Program Grant Number JPMJAP2303, and the Value Exchange Engineering, a joint research project between Mercari, Inc.
and RIISE.
\end{acks}

\bibliographystyle{ACM-Reference-Format}
\bibliography{egbib}
\appendix

\renewcommand{\qualfrow}[3]{
        \resizebox{0.31\textwidth}{!}{%
        \begin{tabular}{@{}c@{\,\,}c@{}}
        & 
        \includegraphics[width=1.85\qualfw]{figure/zoom_results/#1_#2_patch_box.jpg} 
        \end{tabular}
        \begin{tabular}{@{}c@{\,\,}c@{c}}
        
        \includegraphics[width=0.8\qualfw]{figure/zoom_results/#1_#2_patch.jpg}\\
    \includegraphics[width=0.8\qualfw]{figure/zoom_results/#1_#2_patch_2.jpg}\\ 
    
    \end{tabular}
    }
}

\newcommand{\qualfrowe}[3]{
        \resizebox{0.31\textwidth}{!}{%
        \begin{tabular}{@{}c@{\,\,}c@{}}
        & 
        \includegraphics[width=1.8\qualfw]{figure/zoom_results/#1_#2_patch_box.jpg} 
        \end{tabular}
        \begin{tabular}{@{}c@{\,\,}c@{c}}
        
        \includegraphics[width=0.88\qualfw]{figure/zoom_results/#1_#2_patch.jpg}\\
    \includegraphics[width=0.88\qualfw]{figure/zoom_results/#1_#2_patch_2.jpg}\\ 
    
    \end{tabular}
    }
}

\begin{figure*}
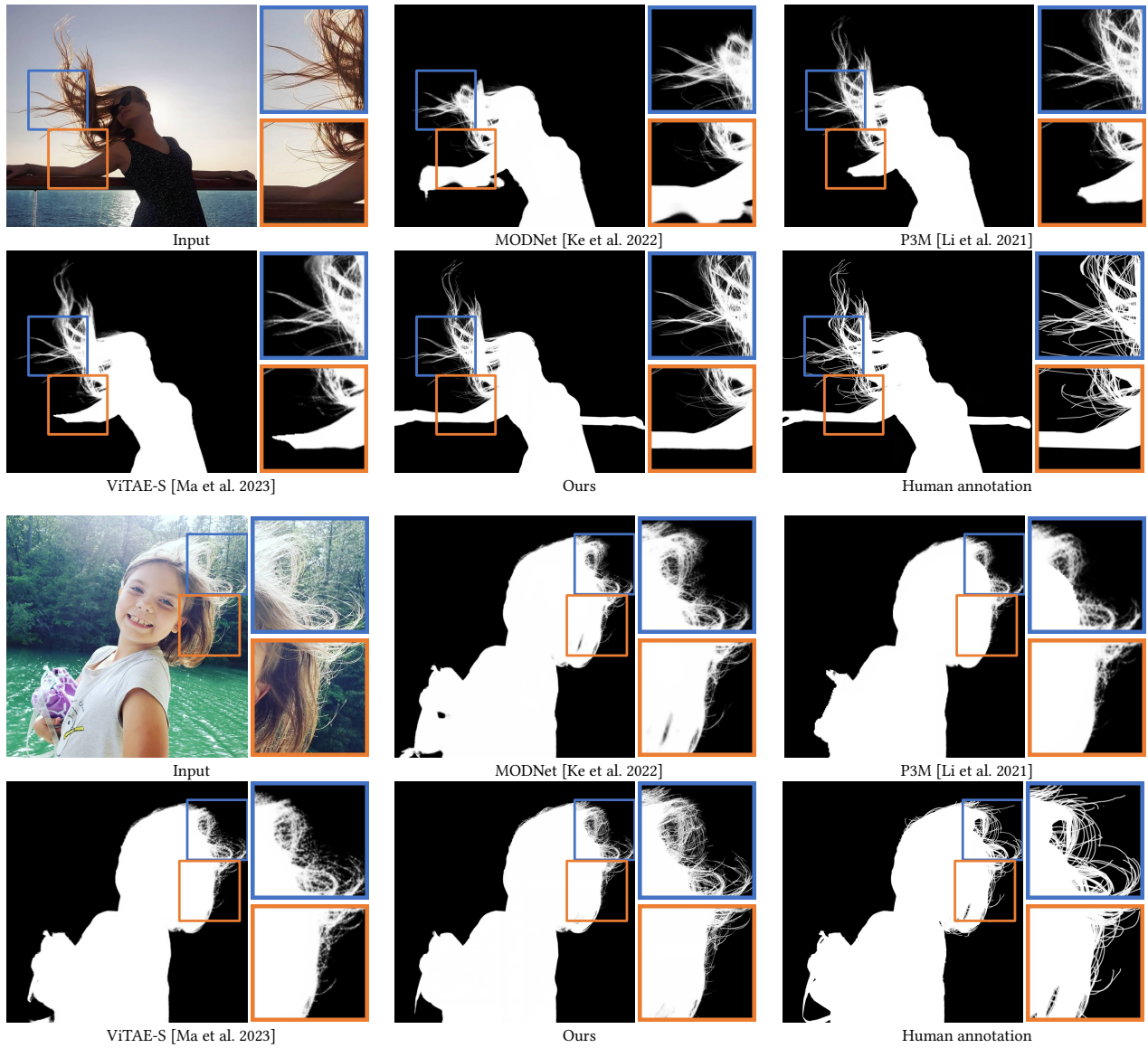

    \centering
    \footnotesize
     \begin{tabular}
    {
        @{\hspace{0mm}}c@{\hspace{1.2mm}} 
        @{\hspace{0mm}}c@{\hspace{1.2mm}}
        @{\hspace{0mm}}c@{\hspace{1.2mm}}
        @{\hspace{0mm}}c@{\hspace{1.2mm}}
        @{\hspace{0mm}}c@{\hspace{1.2mm}} 
        @{\hspace{0mm}}c@{\hspace{1.2mm}} 
    }

    \qualfrow{01277_input}{img}{}
    & \qualfrow{01277_input}{modnet}{}
    & \qualfrow{01277_input}{p3m}{}\\
    Input & MODNet~\cite{Ke-2022-MODNet} & P3M~\cite{Li-2021-P3M} \\
    \qualfrow{01277_input}{p3m_net}{} &   \qualfrow{01277_input}{ours}{} & 
    \qualfrow{01277_input}{gt}{} 
    \\
    ViTAE-S~\cite{MA-2023-P3M-ViTAE} & Ours & Human annotation \\
    \\
    \qualfrowe{02254_input}{img}{}
    & \qualfrowe{02254_input}{modnet}{}
    & \qualfrowe{02254_input}{p3m}{}\\
    Input & MODNet~\cite{Ke-2022-MODNet} & P3M~\cite{Li-2021-P3M} \\
    \qualfrowe{02254_input}{p3m_net}{} &   \qualfrowe{02254_input}{ours}{} & 
    \qualfrowe{02254_input}{gt}{} 
    \\
    ViTAE-S~\cite{MA-2023-P3M-ViTAE} & Ours & Human annotation \\
    \end{tabular}
    \caption{Visual results of guidance-free matting on RVP~\cite{Yu-2021-MG} dataset.}
    \label{fig:visual2}
    \Description{Visual results of guidance-free matting on RVP.}
\end{figure*}

\setlength{\qualw}{0.48\columnwidth}
\renewcommand{\qualrow}[5]{            
        {\includegraphics[trim={#2 #3 #4 #5}, clip,width=\qualw]{figure/blur_ex#1_input.jpg}} &
        {\includegraphics[trim={#2 #3 #4 #5}, clip,width=\qualw]{figure/blur_ex#1_p3mnet.jpg}} &
        {\includegraphics[trim={#2 #3 #4 #5}, clip,width=\qualw]{figure/blur_ex#1_ours.jpg}} &
        {\includegraphics[trim={#2 #3 #4 #5}, clip,width=\qualw]{figure/blur_ex#1_gt.jpg}} 
}
\begin{figure*}[!p]
    \footnotesize
    \centering
    \begin{tabular}
    {   
        @{\hspace{0mm}}c@{\hspace{1.2mm}} 
        @{\hspace{0mm}}c@{\hspace{1.2mm}}
        @{\hspace{0mm}}c@{\hspace{1.2mm}}
        @{\hspace{0mm}}c@{\hspace{1.2mm}} 
    }
         \qualrow{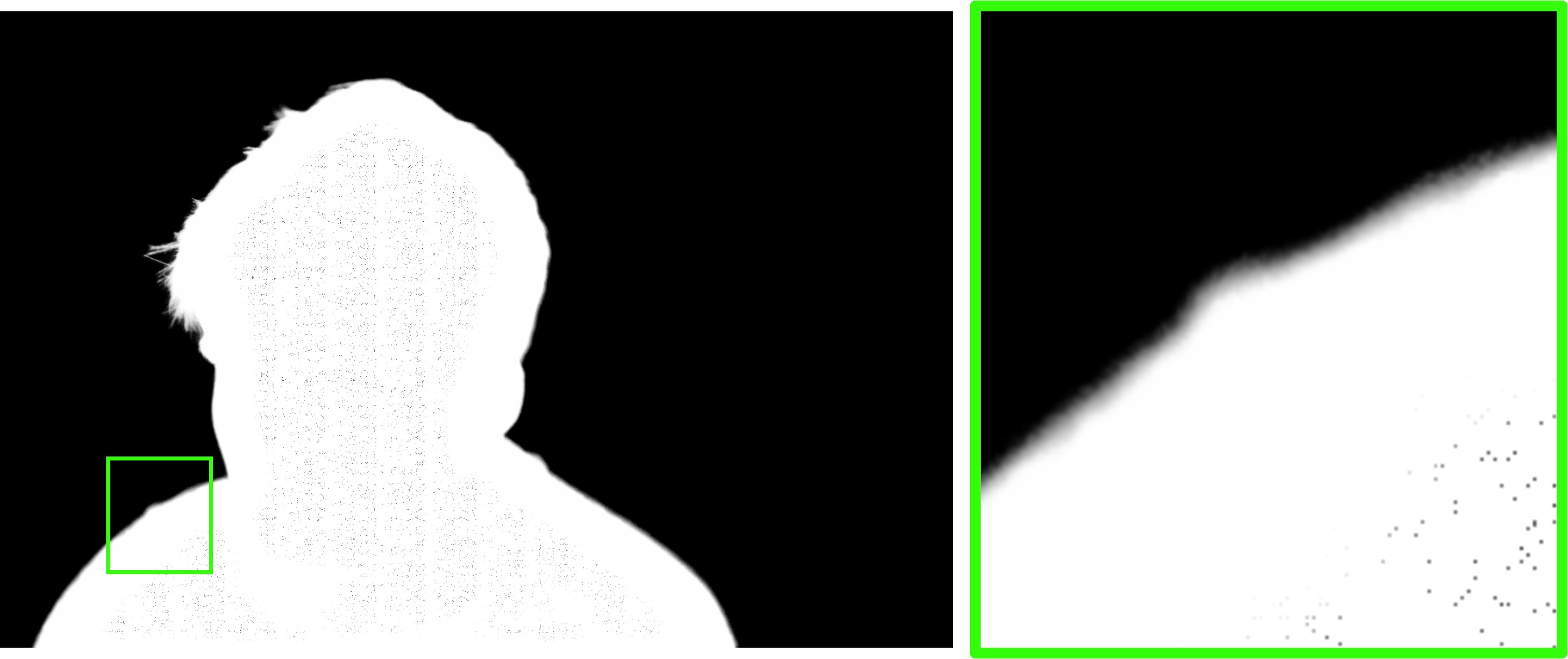}{0}{0}{0}{0}\\
         \qualrow{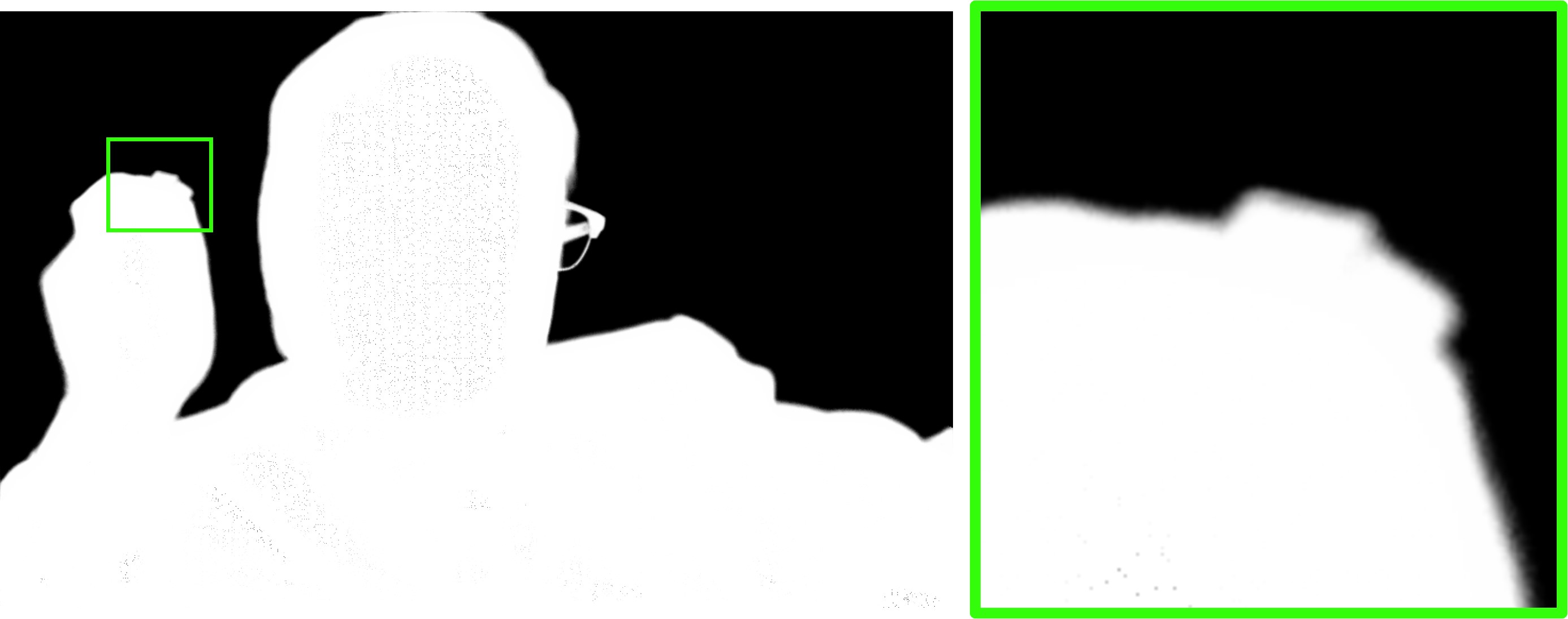}{0}{0}{0}{0}\\
         Input &  ViTAE-S~\cite{MA-2023-P3M-ViTAE} & Ours & Human annotation
    \end{tabular}
    \caption{Matting with out-of-focus blur. Compared to the hard label in the out-of-focus regions of the human annotations, we generate soft mattes.}
    \label{fig:blur}
    \Description{Matting with out-of-focus blur.}
\end{figure*}

\setlength{\qualw}{0.3\textwidth}
\renewcommand{\qualrow}[5]{            
	{\includegraphics[trim={#2 #3 #4 #5}, clip,width=\qualw]{figure/main_dmvsours/#1_diffmat.jpg}} &
        {\includegraphics[trim={#2 #3 #4 #5}, clip,width=\qualw]{figure/main_dmvsours/#1_ours.jpg}} &
        {\includegraphics[trim={#2 #3 #4 #5}, clip,width=\qualw]{figure/main_dmvsours/#1_gt.jpg}} &
        
}
\begin{figure*}
    \centering
    \footnotesize
     \begin{tabular}
    {
        @{\hspace{0mm}}c@{\hspace{1.2mm}} 
        @{\hspace{0mm}}c@{\hspace{1.2mm}}
        @{\hspace{0mm}}c@{\hspace{1.2mm}}
        @{\hspace{0mm}}c@{\hspace{1.2mm}}
    }
    DiffMat~\cite{xu2023diffusionmat} & Ours & Human annotation \\ 
    MSE: 1.0 / SAD: 5.2 & MSE: 1.5 / SAD: 5.4 & -- / -- \\
    \qualrow{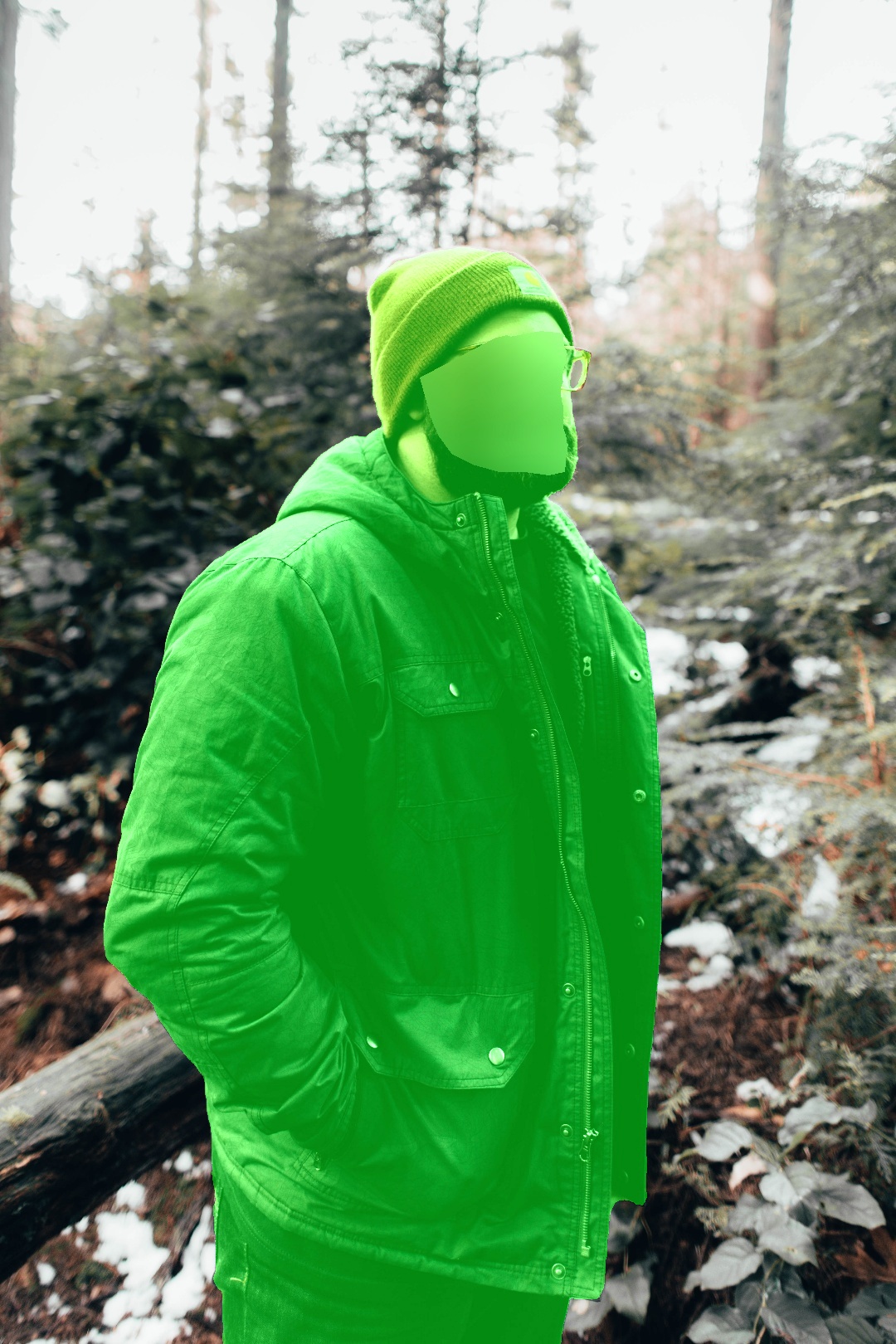}{0cm}{0cm}{0cm}{10cm}\\
    \qualrow{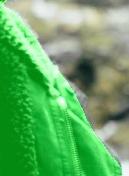}{0cm}{0cm}{0cm}{0cm}
    \end{tabular}
    \vspace{-2mm}
    \caption{Results of trimap-based matting. Our visual results look better, but our evaluation score is worse than DiffMat, mainly because of the imperfect human annotation.}
    \label{fig:guidancematting}
    \Description{Results of trimap-based matting.}
\end{figure*}

\setlength{\qualw}{0.38\columnwidth}
\renewcommand{\qualrow}[5]{            
        {\includegraphics[trim={#2 #3 #4 #5}, clip,width=\qualw]{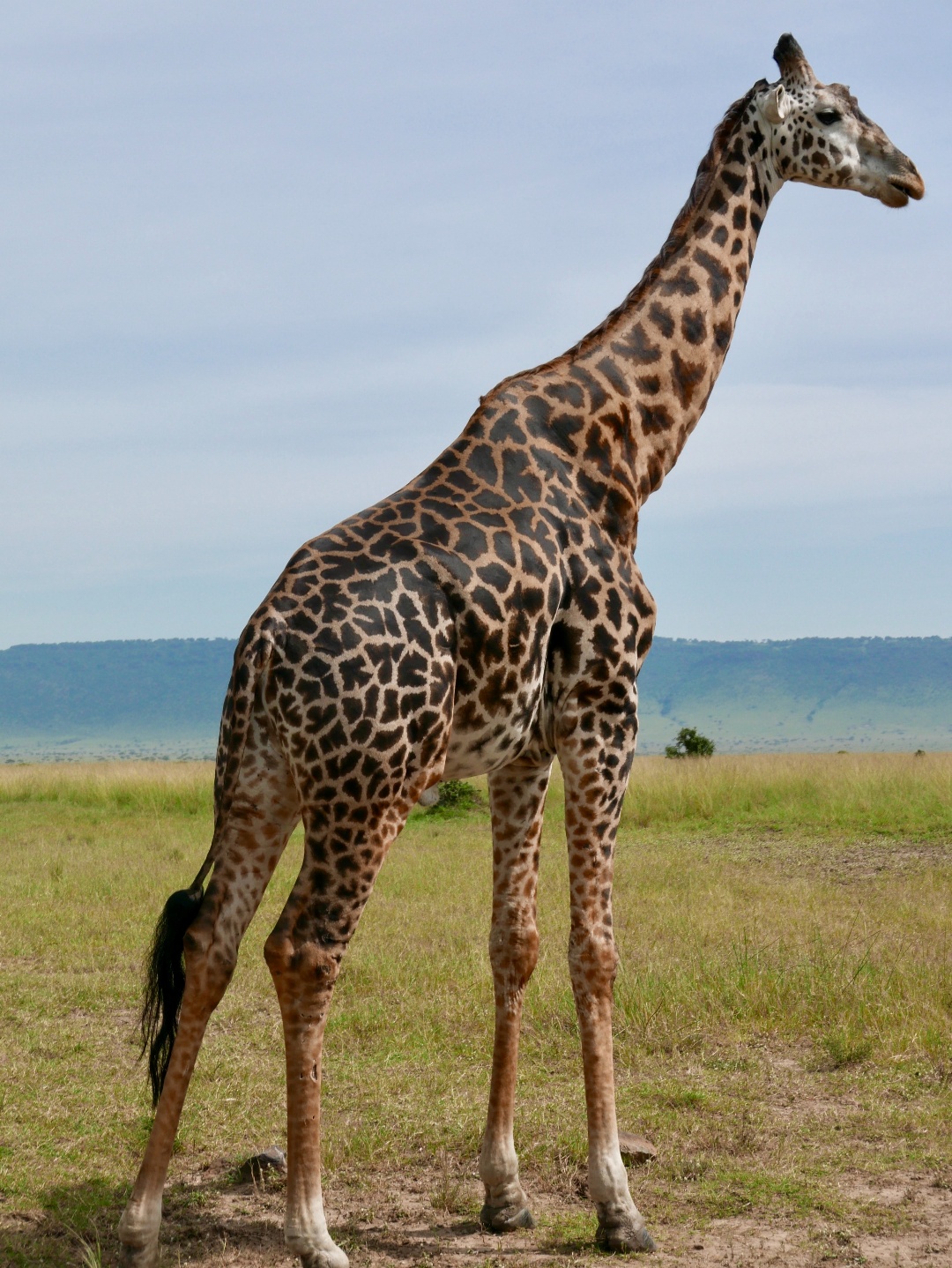}} &
        {\includegraphics[trim={#2 #3 #4 #5}, clip,width=\qualw]{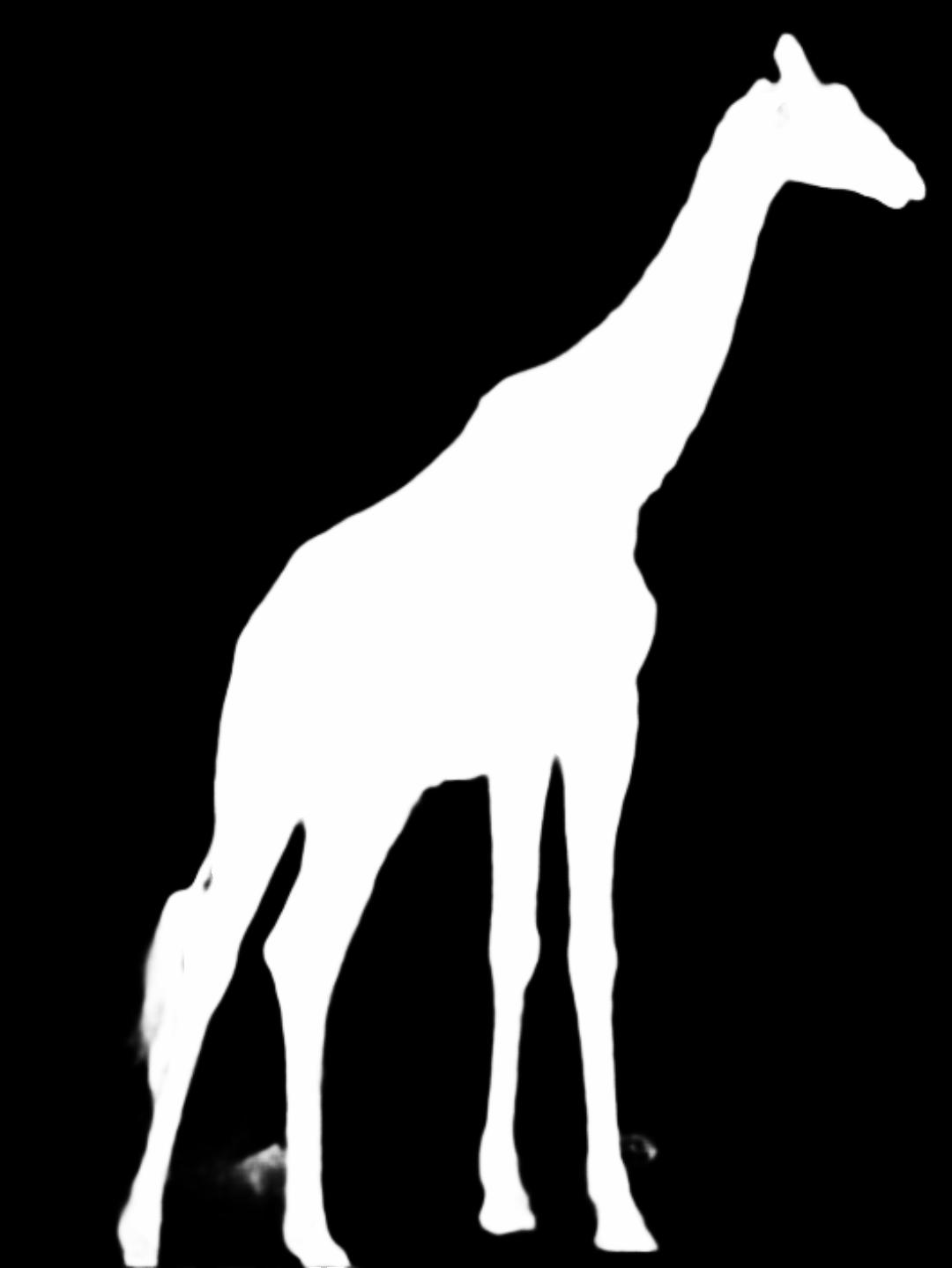}} &
        {\includegraphics[trim={#2 #3 #4 #5}, clip,width=\qualw]{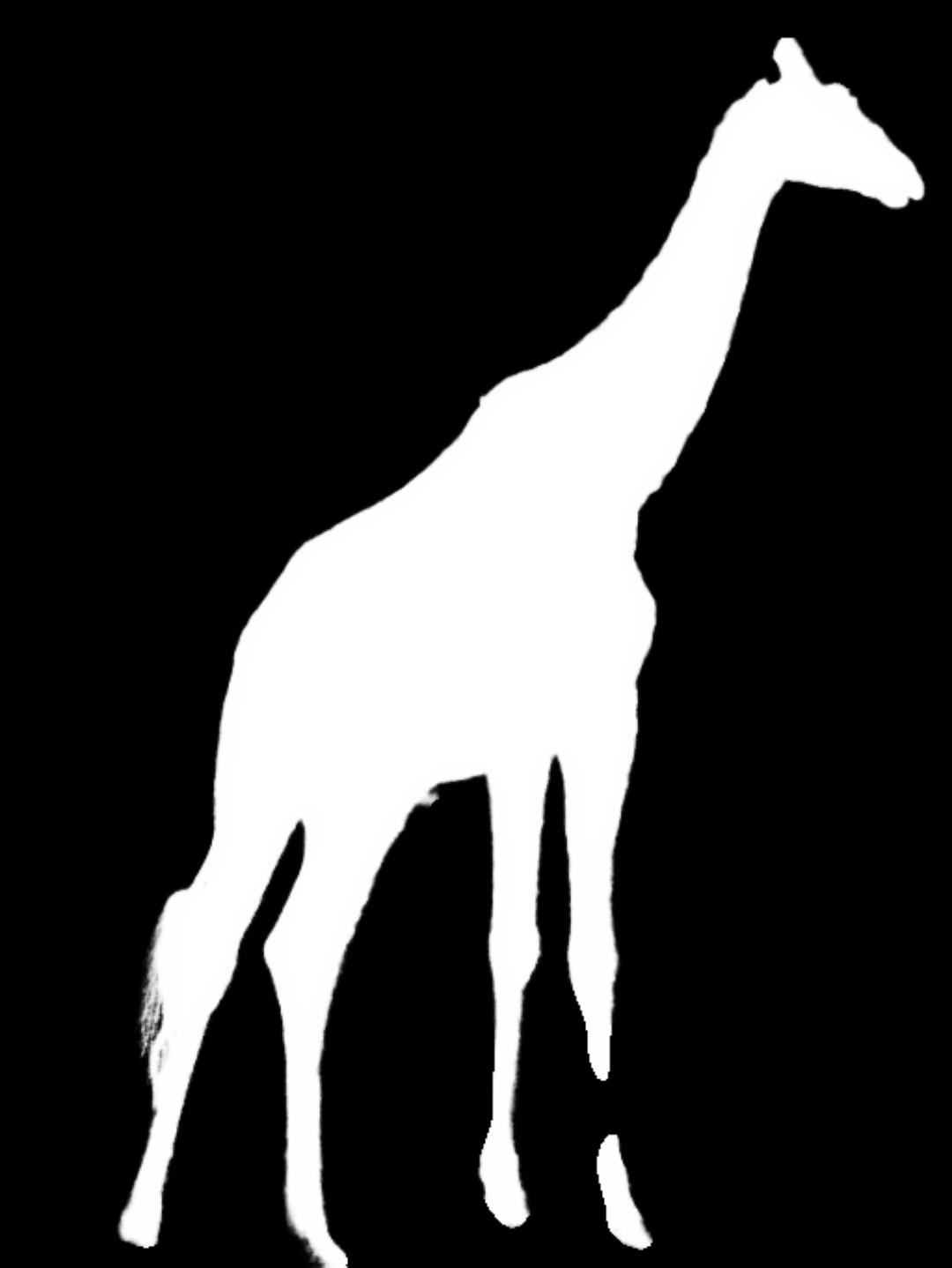}}
        & {\includegraphics[trim={#2 #3 #4 #5}, clip,width=\qualw]{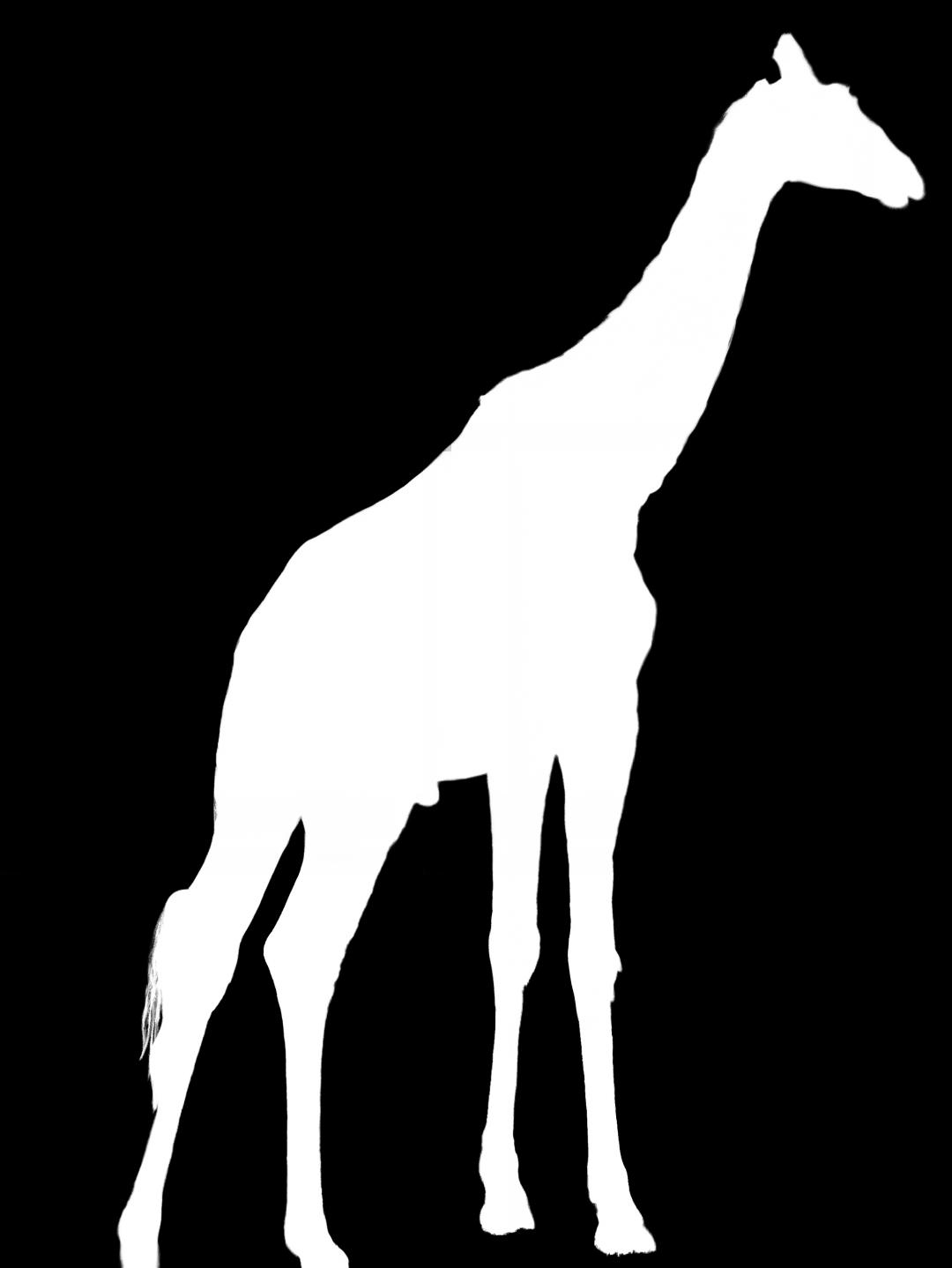}} & {\includegraphics[trim={#2 #3 #4 #5}, clip,width=\qualw]{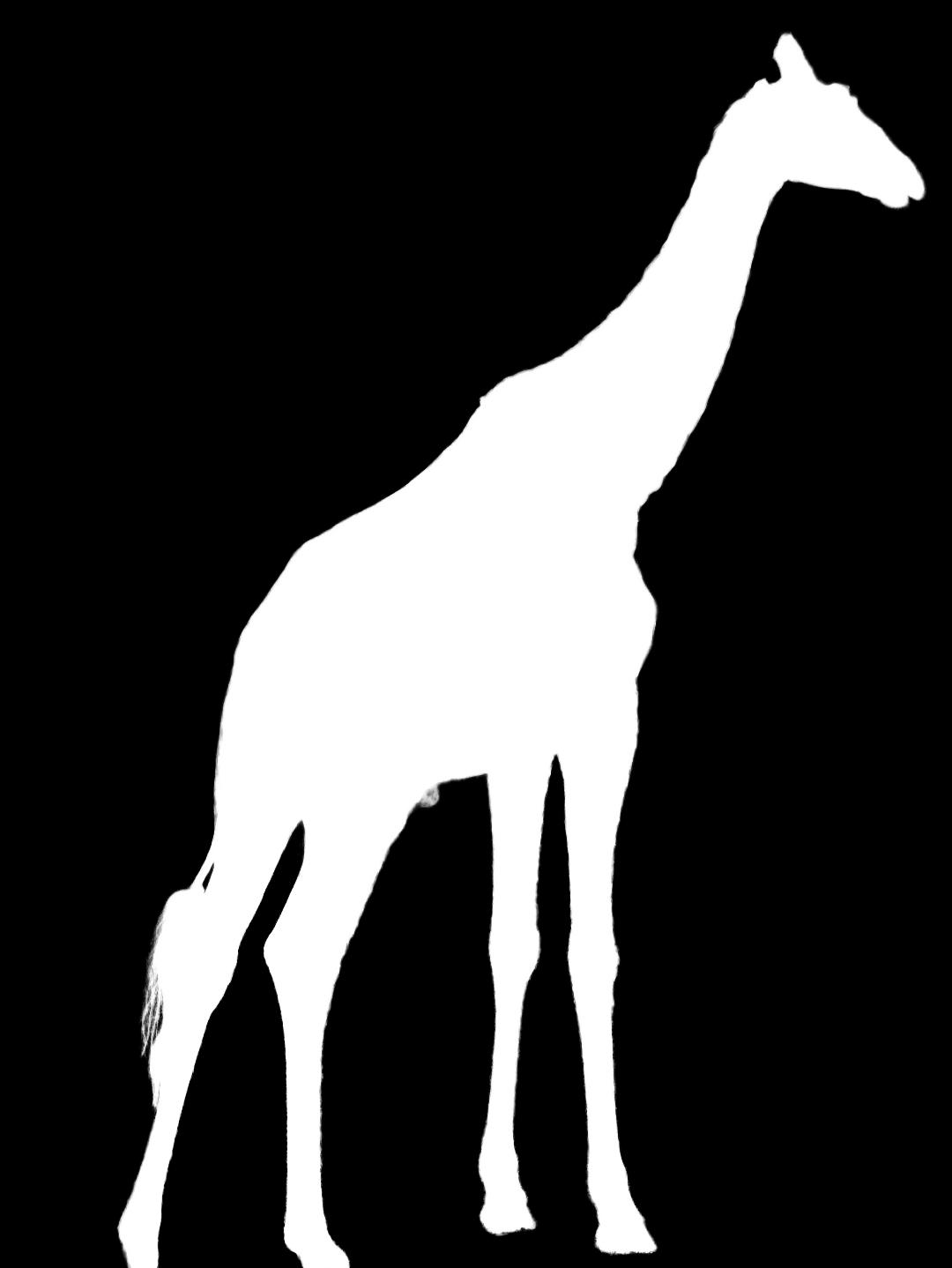}}\\
        Input & MAM~\cite{Li-2023-MattingAny} & ViTAE-S~\cite{MA-2023-P3M-ViTAE} & Ours & Human annotation
}
\begin{figure*}[!p]
    \footnotesize
    \centering
    \begin{tabular}
    {   
        @{\hspace{0mm}}c@{\hspace{1.2mm}} 
        @{\hspace{0mm}}c@{\hspace{1.2mm}}
        @{\hspace{0mm}}c@{\hspace{1.2mm}}
        @{\hspace{0mm}}c@{\hspace{1.2mm}} 
         @{\hspace{0mm}}c@{\hspace{1.2mm}} 
    }
         \qualrow{p_4921a2ba}{0}{0}{0}{0}
    \end{tabular}
    \vspace{-2mm}
    \caption{Matting beyond portraits. Based on SAM, MAM can generate a semantically correct alpha matte for the giraffe image but sacrifice some detail. ViTAE-S, on the other hand, fails to produce a semantically correct result and loses details. Our result closely matches the human annotation.
    }
    \label{fig:beyound}
    \Description{Matting beyond portraits.}
\end{figure*}

\end{document}